\PassOptionsToPackage{table,xcdraw}{xcolor}   % 解决 rowcolor
\PassOptionsToPackage{numbers}{natbib}        % 解决 \citep

\documentclass[sn-basic,iicol]{sn-jnl}

\newcommand{\greenarrow}{\textcolor{green!50!black}{$\uparrow$}}
\usepackage{graphicx}%
\usepackage{multirow}%
\usepackage{amsmath,amssymb,amsfonts}%
\usepackage{amsthm}%
\usepackage[title]{appendix}%
\usepackage{textcomp}%
\usepackage{manyfoot}%
\usepackage{booktabs}%
\usepackage{algorithm}%
\usepackage{algorithmicx}%
\usepackage{algpseudocode}%
\usepackage{listings}%
\usepackage{tabularx}
\theoremstyle{thmstyleone}%
\theoremstyle{thmstyletwo}%

\theoremstyle{thmstylethree}%

\usepackage[table]{xcolor}
\usepackage{tikz}
\usepackage{pifont}
\usepackage[T1]{fontenc}
\usepackage{booktabs}
\usepackage{multirow}
\usepackage{setspace}
\usepackage{graphicx}
\usepackage{subcaption}
\usepackage{float}
\usepackage{array} 
\usepackage{adjustbox} % 控制旋转+缩放

\begin{document}

\title[Article Title]{ContextFusion and Bootstrap: An Effective Approach to Improve Slot Attention-Based Object-Centric Learning}

%%=============================================================%%
%% GivenName	-> \fnm{Joergen W.}
%% Particle	-> \spfx{van der} -> surname prefix
%% FamilyName	-> \sur{Ploeg}
%% Suffix	-> \sfx{IV}
%% \author*[1,2]{\fnm{Joergen W.} \spfx{van der} \sur{Ploeg} 
%%  \sfx{IV}}\email{iauthor@gmail.com}
%%=============================================================%%

\author[1]{\fnm{Pinzhuo} \sur{Tian}}\email{pinzhuo@shu.edu.cn}

\author[1]{\fnm{Shengjie} \sur{Yang}}\email{sjyang@shu.edu.cn}

\author[1]{\fnm{Hang} \sur{Yu}}\email{yuhang@shu.edu.cn}

\author[2]{\fnm{Alex C.} \sur{Kot}}\email{eackot@ntu.edu.sg}

\affil[1]{\orgdiv{School of Computer Engineering and Science}, \orgname{Shanghai University}, \city{Shanghai}, \country{China}}

\affil[2]{\orgdiv{School of Electrical and Electronic Engineering}, \orgname{Nanyang Technological University}, \country{Singapore}}

%%==================================%%
%% Sample for unstructured abstract %%
%%==================================%%

\abstract{A key human ability is to decompose a scene into distinct objects and use their relationships to understand the environment. Object-centric learning aims to mimic this process in an unsupervised manner. Recently, the slot attention-based framework has emerged as a leading approach in this area and has been widely used in various downstream tasks. However, existing slot attention methods face two key limitations: (1) a lack of high-level semantic information. In current methods, image areas are assigned to slots based on low-level features such as color and texture. This makes the model overly sensitive to low-level features and limits its understanding of object contours, shapes, or other semantic characteristics. (2) The inability to fine-tune the encoder. Current methods require a stable feature space throughout training to enable reconstruction from slots, which restricts the flexibility needed for effective object-centric learning. To address these limitations, we propose a novel ContextFusion stage and a Bootstrap Branch, both of which can be seamlessly integrated into existing slot attention models. In the ContextFusion stage, we exploit semantic information from the foreground and background, incorporating an auxiliary indicator that provides additional contextual cues about them to enrich the semantic content beyond low-level features. In the Bootstrap Branch, we decouple feature adaptation from the original reconstruction phase and introduce a bootstrap strategy to train a feature-adaptive mechanism, allowing for more flexible adaptation. Experimental results show that our method significantly improves the performance of different SOTA slot attention models on both simulated and real-world datasets.}


%%\pacs[JEL Classification]{D8, H51}

%%\pacs[MSC Classification]{35A01, 65L10, 65L12, 65L20, 65L70}

\maketitle

\section{Introduction}\label{sec1}

In perceiving the surrounding environment, humans typically divide it into distinct objects, subsequently understanding the scene through the relationships and connections among these objects \citep{marcus2003algebraic,lake2017building}. Thus, enabling our model to decompose the visual scene into different objects is highly meaningful and lays the foundation for reasoning \citep{mansouri2023object}. Unsupervised object-centric learning \citep{greff2020binding} is an approach designed to address this challenge, aiming to represent a scene as a composition of distinct objects using only visual information.

\begin{figure}[t]
    \centering
    \captionsetup[sub]{labelformat=parens, justification=centering} 

    \begin{subfigure}[b]{\linewidth}  
        \centering
        \includegraphics[width=\linewidth]{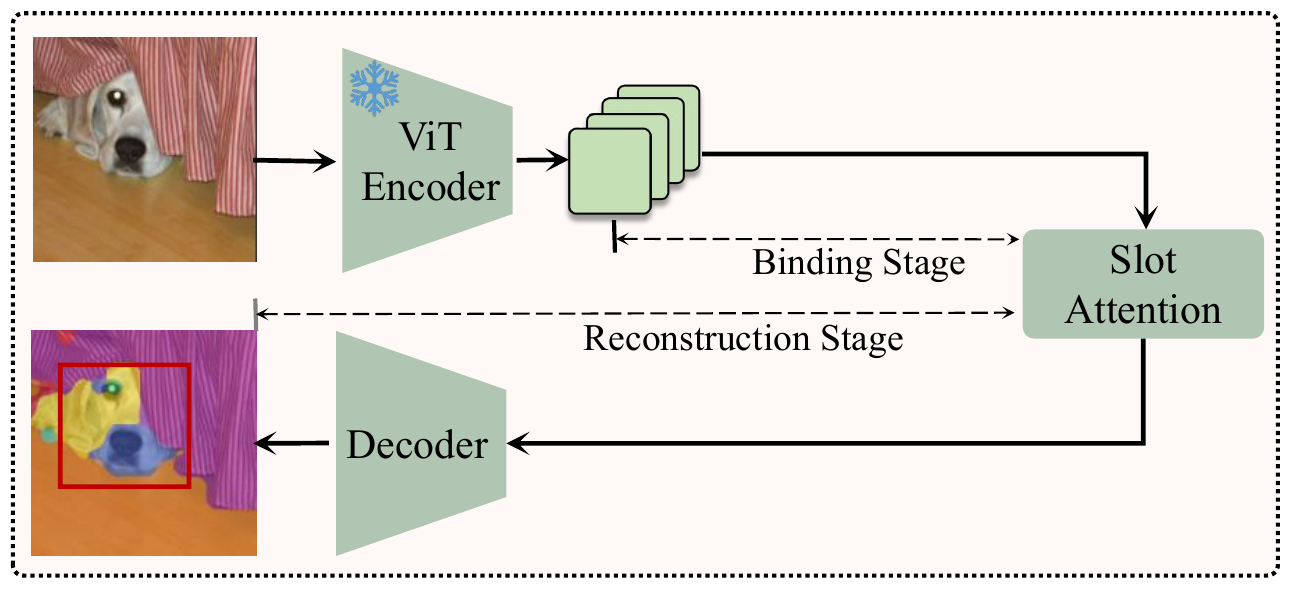}
        \caption{Classical slot attention-based object-centric learning framework}
        \label{fig:sub1}
    \end{subfigure}

    \vspace{0.5em}  

    \begin{subfigure}[b]{\linewidth}
        \centering
        \includegraphics[width=\linewidth]{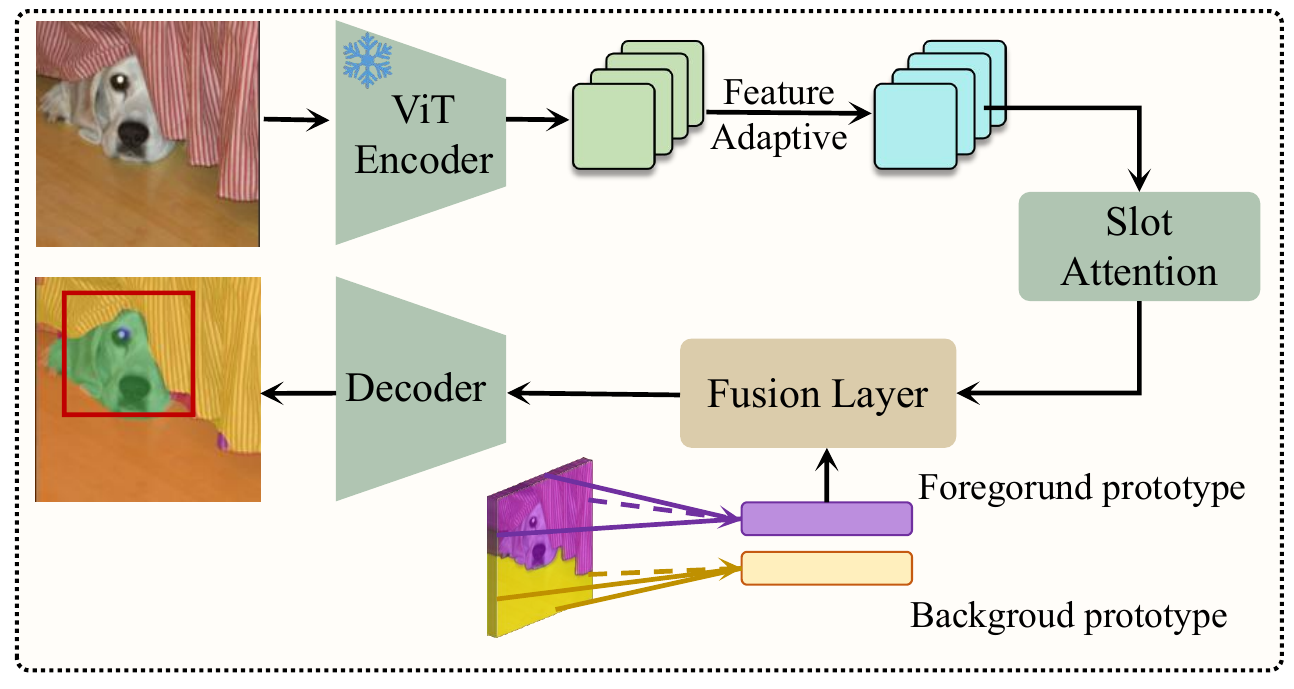}
        \caption{Integrated with our method}
        \label{fig:sub2}
    \end{subfigure}

    \caption{\textbf{The pipelines of the original slot attention-based object-centric learning framework and our proposed method.} Our method can be seamlessly integrated into the existing framework, enhancing its performance by introducing a feature-adaptive layer and incorporating additional semantic information on foreground and background.}
    \label{fig:introduction}
\end{figure}

After significant development, the slot attention-based method \citep{SA} has become the mainstream framework in the field. This kind of method can be divided into two primary stages, as shown in Fig. \ref{fig:introduction} (a): the binding stage and the reconstruction stage. In the binding stage, an attention-based mechanism is used to assign different areas of the image to distinct slots. The reconstruction stage then involves rebuilding the input based on information derived from these slots, ensuring that each slot accurately represents an individual object within the image. These two stages enable the slot attention-based method to achieve impressive results in various areas, such as \citep{qian2023semantics,zheng2024survey}.

However, this framework has two main limitations: \textbf{(1) Lack of High-Level Semantic Information}: the framework relies solely on reconstruction loss for optimization, without incorporating semantic knowledge. As a result, the model is limited to focusing on low-level features, such as color and texture, to cluster pixels into different slots and reconstruct the image. In complex scenes, semantic information becomes crucial. For example, background variations in color or texture might be more prominent and easier for the model to detect than foreground objects. This leads to misleading slot assignments, where the model divides the background into different regions based on low-level cues, struggling to bind slots to foreground objects. \textbf{(2) Inability to Fine-Tune the Encoder}: in current slot attention-based methods, the encoder provides a fixed feature space for clustering image regions into different slots. Many state-of-the-art methods \citep{randomframes} also use this fixed feature space to assess the performance of feature reconstruction derived from the slots and the original features by reconstruction loss. Maintaining a stable feature space across training phases is crucial for preventing clustering collapse. However, pretrained models used as encoders are typically trained by different tasks, such as classification or unsupervised contrastive learning, and may not provide the suitable features needed for pixel-level clustering.

To address these issues, we propose a ContextFusion Stage and a Bootstrap Branch, which collaboratively overcome both limitations. In the ContextFusion stage, we focus on constructing semantic information from the `foreground and background' perspective and integrating it into the slot attention framework. Compared to directly obtaining individual object-level semantic information, foreground and background semantic information are easier to extract and implement based on the semantic cues retained in the encoder. Moreover, the additional foreground and background semantic information can help the model identify objects in the foreground, allowing it to focus more on binding foreground objects while being less influenced by color or texture variations in the background. Specifically, we design an unsupervised auxiliary indicator that learns to differentiate between foreground and background. We then use a lightweight cross-attention layer to fuse this information into the original slot attention pipeline, enriching the model with additional semantic context. Our auxiliary indicator is based on a contrastive learning framework \citep{jaiswal2020survey} and consists of three components: a pixel-level contrastive loss that ensures consistency in pixel predictions across different views, promoting meaningful representation learning; a stuff-level contrastive loss that minimizes variations within the background, reducing misguidance caused by color and texture differences during slot binding; and an entropy-based regularization that strengthens the distinction between foreground and background elements. Finally, the extra foreground and background information are fused into the slot attention framework via the cross-attention layer.

To mitigate the impact of modifying the encoder in the slot attention method, we introduce an additional Bootstrap Branch to decouple feature adaptation from the reconstruction stage. This branch contains a feature-adaptive layer that adapts the encoder’s features, along with a slot attention module to obtain an object mask. For training, we design a bootstrap mechanism that uses the predicted semantic mask from the ContextFusion stage as the target. A matching-based segmentation loss is applied to align each predicted region in the Bootstrap Branch with the corresponding area mask from the ContextFusion stage, replacing the typical reconstruction loss used in slot attention-based frameworks. To ensure that the feature adaptation layer enhances the feature representation capability, we initialize the slot attention module in this branch from scratch and then optimize it using a Weighted Moving Average approach, leveraging the slot attention module from the ContextFusion stage. The adapted features, combined with coarse slots, are expected to achieve the performance improvements observed in the ContextFusion stage. At the testing stage, we combine the feature-adaptive layer with the latest slot attention module fused with the semantic information in the ContextFusion stage to obtain the final performance. However, these components reside in two separate branches. Our design uses a feature adaptation layer to slightly adjust the encoder’s features, rather than fine-tuning the entire encoder. The Bootstrap Branch follows a similar pipeline, using the slot attention module to obtain results. These ensure that the feature after the adaptation layer in the Bootstrap Branch is compatible with the latest slot attention block in the ContextFusion stage. Fig.~\ref{fig:introduction} (b) provides an overview of our method integrated into the existing slot attention-based object-centric learning framework. 

To comprehensively investigate our method, we select two prominent object-centric learning methods, all based on the slot attention framework: DINOSAUR \citep{randomframes} and SPOT \citep{spot}. DINOSAUR introduces a Vision Transformer encoder, making it the first object-centric method applicable to natural images. SPOT leverages a self-training mechanism to provide supervisory signals and enhances the autoregressive decoder by incorporating sequence permutations, achieving the state-of-the-art performance. These methods address the object-centric learning problem from both encoder and decoder perspectives, providing a comprehensive evaluation of our method in varied situations. The results demonstrate that our approach significantly improves their performance on both simulated and real-world datasets. Additional in-depth discussion further emphasizes the effectiveness of the different components of our method.

In summary, our contributions are threefold:
\begin{itemize}
    \item We consider two common issues in the slot attention-based object-centric learning framework: the lack of high-level semantic information and the inability to fine-tune the encoder. A unified method is proposed to address both limitations, which can be integrated with various state-of-the-art models.
    \item To inject more semantic information, we introduce an unsupervised foreground and background auxiliary indicator, which provides additional semantic context to the original slot attention framework. We also propose a novel bootstrap branch to decouple feature adaptation from the reconstruction stage.
    \item Our experimental results demonstrate that integrating our method with various state-of-the-art models significantly enhances their performance both on simulated data and real-world datasets.
\end{itemize}

This article is organized as follows. Section~\ref{sec:related_work}  introduces some related works. Section~\ref{sec:method} describes our method in detail. Section~\ref{sec:experiments} discusses the experiment results conducted on some commonly used datasets. Section~\ref{sec:conclusion} concludes this article.

\section{Related Work}\label{sec:related_work}

\subsection{Object-centric Learning.}
Object-centric learning aims to represent a scene as a composition of multiple independent objects. An early work in this field, IODINE~\citep{IODINE}, utilizes iterative variational inference to represent images via a set of latent variables, each of which describes a distinct object. Similarly, MONet~\citep{monet} jointly trains a segmentation network alongside a variational autoencoder \citep{kingma2013auto} to decompose complex scenes into object-level representation. However, both of these methods rely on multiple encoding-decoding stages, leading to limited computational efficiency. Slot Attention~\citep{SA} addresses this limitation by introducing a simpler iterative attention mechanism. Due to its improved efficiency and effectiveness, the slot-attention-based framework has gradually become the mainstream paradigm in object-centric learning. Nonetheless, despite its success, this method still encounters several challenges when applied to real-world scenarios.

To make the slot attention-based framework more suitable for real-world object-centric scenarios, one direction is to improve the encoder's capabilities. For example, DINOSAUR \citep{randomframes} uses a Vision Transformer \citep{vit} as the encoder and adopts self-supervised pretrained features as reconstruction targets, achieving effective and robust object decomposition in complex real-world scenes. \citep{greff2019multi, kim2024bootstrapping, OCEBO} consider that existing object-centric representation methods, without adjustments to the encoder, limit object discovery capabilities. To address this, they propose task-specific fine-tuning strategies for the encoder, significantly enhancing DINOSAUR’s performance in object discovery and zero-shot generalization in real-world scenarios. Another direction is to improve the accuracy and stability of the slot attention module, through methods such as bi-level optimization \citep{bi1, bi2}, architectural improvements \citep{mo1, mo2}, and the introduction of regularization losses \citep{reg}. Other studies focus on designing more effective decoders to improve scene decomposition capabilities. For instance, SLATE \citep{SLATE} and SlotFormer\citep{wu2022slotformer} employs an autoregressive Transformer decoder, while SPOT \citep{spot} introduces a patch order perturbation strategy to enhance decoding performance. Similarly, methods like \citep{SlotDiffusion, LSD} introduce diffusion-based models, which also show promising results.

In addition to directly improving the performance of the slot attention-based framework, many studies focus on addressing specific limitations when this framework is applied to real-world scenarios. For example, COSA \citep{grounded} binds slots to specific object types, while AdaSlot \citep{adaptive} focuses on dynamically determining the optimal number of slots. These methods mainly target static images, some studies have extended slot-based approaches to video data \citep{video1, video2, video3, video4, video5}, incorporating additional modalities such as optical flow \citep{flow1, flow2} and depth information \citep{depth}, which further enhance object learning performance on complex real-world video data. Against this background, some methods have explored integrating the slot mechanism into specific downstream tasks, like Slot-IVPS \citep{zhou2024object} and SAOA \citep{wang2020symbiotic}.

\subsection{Contrastive Learning.}
Contrastive learning has emerged as a powerful self-supervised learning paradigm, aiming to learn discriminative feature representations by pulling semantically similar samples closer and pushing dissimilar ones apart in the embedding space. Early developments such as Siamese networks \citep{koch2015siamese} and triplet loss \citep{hermans2017defense} laid the foundation for image-level discrimination. The breakthrough came with SimCLR \citep{SimCLR}, which introduced a simple yet effective framework that relies on strong data augmentation, a large batch size, and contrastive loss (NT-Xent) to achieve competitive results in image classification tasks without labels.
With SimCLR, MoCo \citep{MOCO} proposed a momentum encoder and a dynamic memory bank to decouple batch size from the number of negative samples, enabling more scalable training. BYOL \citep{BYOL} and SimSiam \citep{SimSiam} pushed the boundary further by showing that contrastive learning can succeed even without explicit negative samples, relying on architectural asymmetry and stop-gradient mechanisms. Subsequent research also explored multi-modal \citep{CLIP}, and fine-grained contrastive frameworks \citep{SupCon}. 

In addition to image-level contrastive learning, some research has explored pixel-level contrastive learning to help models capture the rich spatial structure within an image for dense prediction tasks. For example, VICRegL \citep{VICRegL} independently regularizes local regions in the feature map, enhancing the consistency and discriminativeness of pixel representations. DetCo \citep{DetCo} combines region proposals with an image-level encoder to contrast object features across scales. SlotCon \citep{slotcon}, based on the slot attention mechanism, enables object grouping and contrast without relying on explicit detection modules, significantly improving representation learning in complex scenes.

\section{Method}\label{sec:method}
\begin{figure*}[t]
    \centering
    \includegraphics[width=0.95\linewidth]{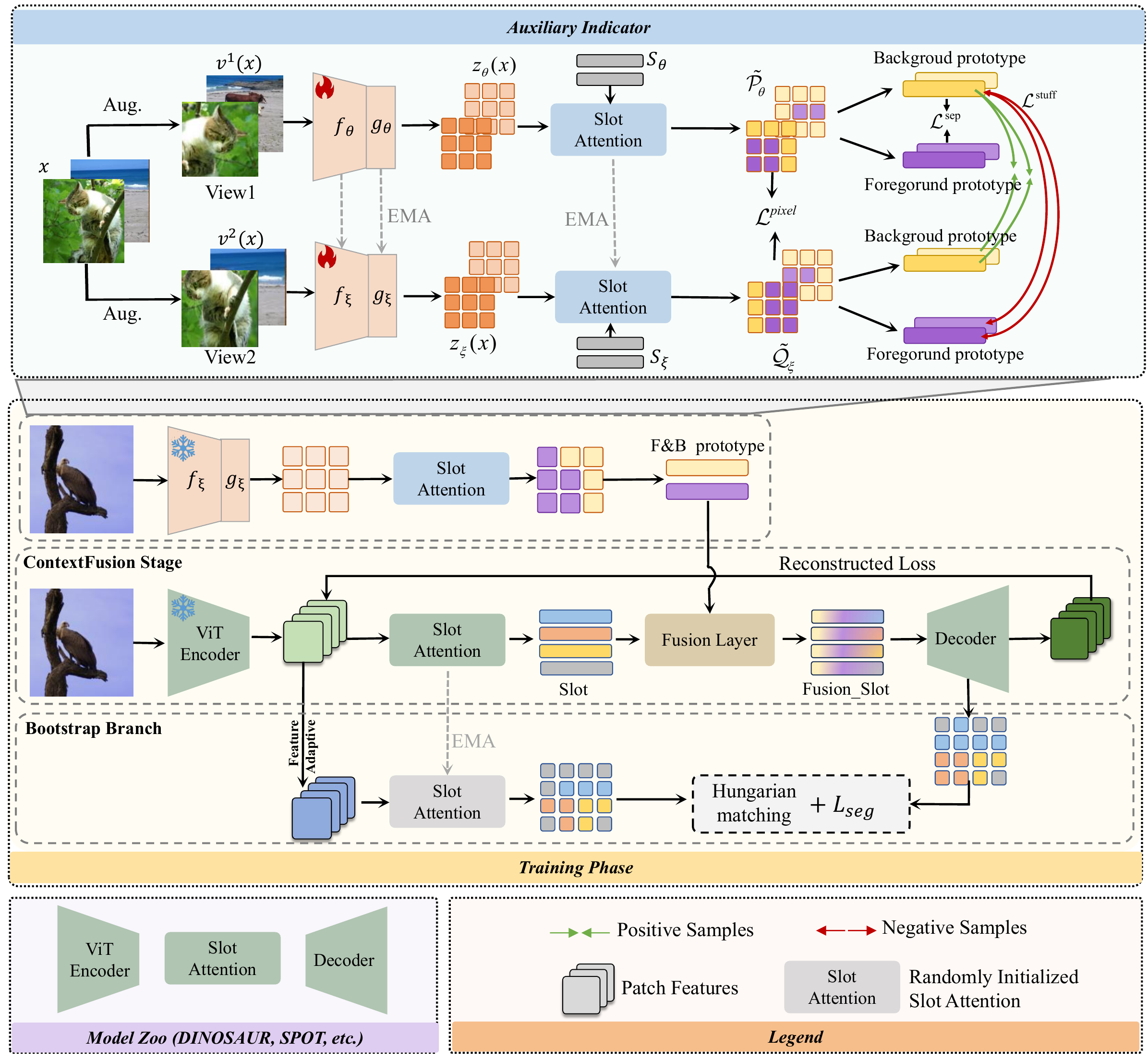}
    \caption{ 
    \textbf{Overview of our proposed method.} An auxiliary indicator is first trained to provide additional semantic information about the foreground and background. Our method takes modules (encoder, slot attention, and decoder) from classical slot attention-based object-centric learning models in the model zoo as input. By integrating extra semantic knowledge in the ContextFusion stage and employing a feature-adaptive layer within the bootstrap branch, our approach aims to enhance the performance of conventional methods.}
    \label{fig:overview method}
\end{figure*}

Our framework consists of two components: 1) The ContextFusion Stage. In this stage, we first train a foreground and background auxiliary indicator to provide additional semantic information about foreground objects and background elements. This extra semantic information is then integrated into the existing slot attention framework through a fusion layer, enhancing overall performance. 2) The Bootstrap Branch: During the feature fusion process in the ContextFusion stage, the predictions from the ContextFusion stage are also used as targets to simultaneously bootstrap the feature adaptation layer to improve the feature representation. Fig. \ref{fig:overview method} shows the overview of our method.

\subsection{ContextFusion Stage}
\textbf{\textit{Auxiliary Indicator.}} The design of our method is driven by two main goals: it must be an unsupervised approach, and it should provide accurate and robust semantic information to distinguish between the foreground and background. Compared to directly provided object-level semantic information, foreground and background region-level semantic information are easier to obtain and also very important. Pre-trained models based on unsupervised contrastive learning, which have been widely used in various fields \citep{kweon2023weakly}, can offer initial semantic cues for foreground objects and background regions. Therefore, we use a pre-trained model as the backbone, instead of training from scratch. In the following, we explain how we leverage the initial semantic information embedded in the pre-trained model to accurately capture semantic knowledge about the foreground and background regions.

The implementation of our indicator leverages the advanced teacher-student contrastive learning framework. Specifically, our approach involves a student network, denoted by $\theta$, and a teacher network, denoted by $\xi$. Both networks share the same architecture but use different sets of weights. The parameters of the teacher network, $\xi$, are updated as an exponential moving average of the student network, $\theta$. In each network, we use two distinct slots, $S = \{s^f, s^b\}$, \textit{i.e.}, $S_{\theta}$ and $S_{\xi}$ for the student and teacher networks, respectively, to help recognize the foreground and background in each image. $S_{\xi}$ is also optimized by EMA.

Unlike the classical slot attention method, $s^f_{\theta~\text{or}~\xi}$ and $s^b_{\theta~\text{or}~\xi}$ are given semantic meaning, enabling the classification of pixels into their corresponding semantic groups, such as foreground and background. Here, $s^f_{\theta}$ and $s^b_{\theta}$ represent the slots for foreground and background in the student network, respectively, while $s^f_{\xi}$ and $s^b_{\xi}$ represent the slots for foreground and background in the teacher network, respectively. 

Our indicator contains three aspects. The first is the pixel-level consistency. For different augmentations of the same image, we ensure that pixels in the same location receive similar assignment scores. Given an input image $x$, two random augmentations are applied to produce two augmented views $v^1(x)$ and $v^2(x)$. Each augmentation view is then encoded with an encoder $f$ into a hidden feature map and then transformed with a multilayer perceptron (MLP) $g$ to get the feature map $z$, \textit{i.e.}, $z_{\theta}(x) = g_\theta(f_{\theta}(v^1(x)))$ and $z_{\xi}(x) = g_\xi(f_{\xi}(v^2(x)))$. The fore-background representation vectors of the image $x$ in different networks can be obtained via slot attention module, \textit{i.e.}, $\hat{S}_{\theta}(x) = \text{Slot Attention}(z_{\theta}(x), S_{\theta})$, $\hat{S}_{\xi}= \text{Slot Attention}(z_{\xi}(x), S_{\xi})$. Then, the pixel assignment matrix, defining the score fore or background each pixel belongs to, $\mathcal{P}_{\theta}$ and $\mathcal{Q}_{\xi}$ can obtain by $\mathcal{P}_{\theta}(x) = z_{\theta}(x) \cdot {\hat{S}_{\theta}}^{\top}$. The assignment matrix $\mathcal{Q}_{\xi} (x)$ for the teacher network can be obtained in the same way, $\mathcal{Q}_{\xi}(x) = z_{\xi}(x) \cdot {\hat{S}_{\xi}}^{\top}$.

To keep the pixel level consistency, we enforce the prediction of the two views assignment $\mathcal{P}_{\theta}(x)$ and $\mathcal{Q}_{\xi}(x)$ match each other. However, the geometric image augmentations, such as random crop, scale, or flip, between different views will cause the scale or layout of the two feature maps to be inconsistent. So similar to \citep{slotcon}, we use the inverse augmentation process on the predicted assignments to align their spatial locations, \textit{i.e.}, $
\widetilde{\mathcal{P}}_\theta (x)=\operatorname{invaug}\left(\mathcal{P}_{\theta}(x)\right), 
\tilde{\mathcal{Q}}_{\xi} (x)=\operatorname{invaug}\left(\mathcal{Q}_{\xi}(x)\right)$.

Based on the aligned assignments, we apply the cross-entropy loss to enforce the consistency in assignment scores between spatially aligned pixels from different views.
The cross-entropy loss is averaged over all spatial locations to keep the pixel prediction consistent, as:
\begin{equation}
\begin{split}
      \mathcal{L}^{\text {pixel}}(x)= &\frac{1}{H \times W} \sum_{i, j}   \mathcal{L}^{\mathrm{CE}}\left({\tilde{\mathcal{Q}}}_{\xi}(x)[i, j], {\tilde{\mathcal{P}}}_\theta(x)[i, j]\right) .  
\end{split}
\end{equation}
$i,j$ means the position in the assignment matrix. $H, W$ are the height and width of the final image feature.

In addition to the pixel-level contrastive loss, we incorporate a stuff-level consistency loss. This loss aims to alleviate variations in the background by closing the background in different images as positive samples, while considering all foreground samples as negative samples. According to the pixel assignment matrix $\tilde{\mathcal{P}}_\theta (x)$ and $\tilde{\mathcal{Q}}_\xi (x)$, we can obtain the prototypes of the foreground and background by region average pooling to represent the whole foreground and background regions of the image $x$. $p_{\theta}^f(x), p_{\theta}^b(x)$ represent the foreground and background prototypes obtained from the student network. $p_{\xi}^f(x), p_{\xi}^b(x)$ are from the teacher network. Then, the stuff-level loss can be written as follows:
% \begin{equation}\label{eq:stuff}
% \begin{split}
%      \mathcal{L}^\text{stuff}(B) =  \frac{1}{B(B-1)}\sum_{i=1}^B \sum_{j\neq i}^B \left[ & \text{sim} (p^{b}_{\theta}(x_i), p^b_{\xi}(x_j)) -  \right. \\
%   &  \left. \text{sim} (p^{b}_{\theta}(x_i), p^{f}_{\xi}(x_j) )
%     \right].\\
% \end{split}
% \end{equation}

\begin{equation}\label{eq:stuff}
\begin{split}
     \mathcal{L}^\text{stuff}(B) = \frac{1}{B(B-1)} \sum_{i=1}^B \sum_{j\neq i}^B \Big[
     & \text{sim}(p^b_{\theta}(x_i), p^b_{\xi}(x_j)) 
      - \\ & \text{sim}(p^b_{\theta}(x_i), p^f_{\xi}(x_j))
     \Big]
\end{split}
\end{equation}

In Eq.~\ref{eq:stuff}, we hope the similarity between the background across different views will be large, however, the similarity between foreground and background should be small. We adopt the cosine similarity in our method. $B$ means the batch of training data.

To encourage the semantic centers to emphasize different visual contents and cover
diverse semantics to avoid collapse, we apply a regularization to the assignment matrix of different views, as follows:
\begin{equation}
    \mathcal{L}^{\text {sep}}(x)= \text{Entroy}({\mathcal{P}_{\theta}}(x)) + \text{Entropy}(\mathcal{Q}_{\xi}(x)).
\end{equation}

The total loss to optimize the proposed indicator is
\begin{equation}\label{eq:loss_all}
    \mathcal{L} = \mathcal{L}^\text{pixel}(x) - \lambda \mathcal{L}^\text{stuff}(B) - \gamma \mathcal{L}^\text{sep}(x).
\end{equation}

\textit{\textbf{Fusion Layer.}} After training the auxiliary indicator, we use the fusion layer to integrate the semantic information provided by the indicator into the slot attention-based models. In the slot attention framework, the learned slots represent different image regions, each corresponding to distinct objects. Our auxiliary indicator provides slots that represent the foreground and background regions with semantic information, indicating which parts belong to the foreground and which parts belong to the background. Therefore, we apply a fusion layer to combine the slots derived from the image with the foreground and background representation vectors to obtain the final slots, completing the image region binding. Because our indicator processes background elements uniformly, ensuring feature consistency across different background areas. By combining the foreground and background slots generated by our method with those produced by existing approaches, the slot-attention based framework is supposed to retain its ability to distinguish between objects while minimizing the impact of the background, resulting in more effective foreground segmentation.

To design the fusion layer, we use a cross-attention layer followed by a self-attention layer, which provides a lightweight and straightforward implementation. The cross-attention layer is employed to establish relationships between the learned slots from the original slot attention method and the foreground and background feature vectors. The self-attention layer, on the other hand, is used to build relationships between the slots for the next reconstruction step.

In the cross-attention layer, foreground and background feature vectors from our auxiliary indicator are used as $K_c$ and $V_c$, respectively. $Q_c$ corresponds to the learned slots from the original slot attention method:
\begin{equation}\label{eq:fusion}
  \text{Output} = \text{Cross-Attention}(Q_c, K_c, V_c), 
\end{equation}
Next, the output sequence from Eq. \ref{eq:fusion} is used as the $Q_s, K_s, V_s$ in the self-attention layer, as follows,
\begin{equation}\label{eq:sft}
\text{Fusion\_slots} = \text{Self-Attention}(Q_s, K_s, V_s).
\end{equation}

The fusion slots are then used as input to the decoder to compute the final reconstruction loss, following the same process as the original slot attention method.

\subsection{Bootstrap Branch}
In both the original slot attention method and our ContextFusion stage, the encoder is not fine-tuned, because it is used to provide a stable feature space for slot binding or to measure the quality of feature reconstruction. Making hasty changes to the encoder can lead to clustering collapse, as highlighted in \citep{FT-DINOSAUR}. To mitigate the impact of modifying the encoder in the slot attention method, we introduce an additional Bootstrap Branch to decouple feature adaptation from the reconstruction branch. During the training of the fusion layer in the ContextFusion stage, the Bootstrap Branch is also trained simultaneously and supposed to provide more appropriate features for slot binding.

In this branch, we incorporate a feature-adaptive layer that modifies the encoder’s features. The feature output by the encoder is passed through the feature-adaptive layer, and the adapted feature is then fed into a slot attention module to generate the predicted object mask for each object region. The design of the feature adaptation layer is simple and lightweight. We apply a linear transformation to the feature obtained from the encoder using additional scale and bias parameters, which can be expressed as:
\begin{equation}\label{eq:fusion layer}
    \hat{z}(x) = \alpha \cdot z(x) + \beta,
\end{equation}
where $x$ is the input image, $z(x)$ is the corresponding feature obtained by the feature encoder. $\alpha$ represents the scale parameter, and $\beta$ is the bias parameter. After linear transformation, the adapted feature $\hat{z}(x)$ is passed into the slot attention module to generate the object mask, following the same process as in the original slot attention methods.
  
To train the feature-adaptive layer (Eq. \ref{eq:fusion layer}) in our Bootstrap Branch, we introduce a bootstrap mechanism that uses the predicted semantic mask obtained from the ContextFusion stage as the target. We then apply a matching-based segmentation loss to align each predicted region from the Bootstrap Branch with the corresponding area in the target mask. This matching-based segmentation loss is similar to \citep{carion2020end}. First, we use the Hungarian algorithm to compute the optimal assignment between the mask predicted by the Bootstrap Branch and the target mask from the ContextFusion stage. Afterward, we use cross-entropy loss to perform pixel classification and finalize the segmentation loss. The full process can be formulated as:
\begin{equation}\label{eq:match-loss}
    \hat{\sigma}=\underset{\sigma \in \mathfrak{S}_N}{\arg \min } \sum_i^N \mathcal{L}_{\text {match}}\left(y_i, \hat{y}_{\sigma(i)}\right),
\end{equation}
Given an image $x$, $y$ is the set of objects predicted by the ContextFusion stage as the target, and $\hat{y} = \{\hat{y}_i\}^N_{i=1}$ is the set of $N$ predicted objects from the Bootstrap Branch. Eq.~\ref{eq:match-loss} is used to search for a permutation of $N$ elements $\sigma \in \mathfrak{S}_N$ with the lowest cost to bipartite matching between $y$ and $\hat{y}$. $\mathcal{L}_{\text {match }}\left(y_i, \hat{y}_{\sigma(i)}\right)$ is a pair-wise matching cost between target object $y_i$ and a prediction with index $\sigma(i)$. This optimal assignment is computed efficiently with the Hungarian algorithm, following previous work \citep{match}. 

After confirming the match by Eq.~\ref{eq:match-loss}, we use the pixel segmentation loss to align the prediction from the Bootstrap Branch to the target, as: 
\begin{equation}\label{eq:segloss}
    \mathcal{L}_{\text {seg}}(x)=\sum_{i=1}^{H\times W}-\log \hat{p}_{\hat{\sigma}(i)}\left(c_i\right),
\end{equation}
where $H \times W$ represents the number of pixels in the prediction of $x$. $c_i$ is the object label of pixel $i$ in the target mask. $\hat{\sigma}(i)$ is the corresponding matching pixel in the predicted mask. Using the mask from the ContextFusion stage as the target and serving as the critic, we allow the feature adaptation layer to be optimized. This approach helps avoid collapse, as it provides an opportunity to refine the feature adaptation layer, unlike relying solely on the feature or patch itself in reconstruction.

\begin{figure}[t]
    \centering
    \includegraphics[width=0.9\linewidth]{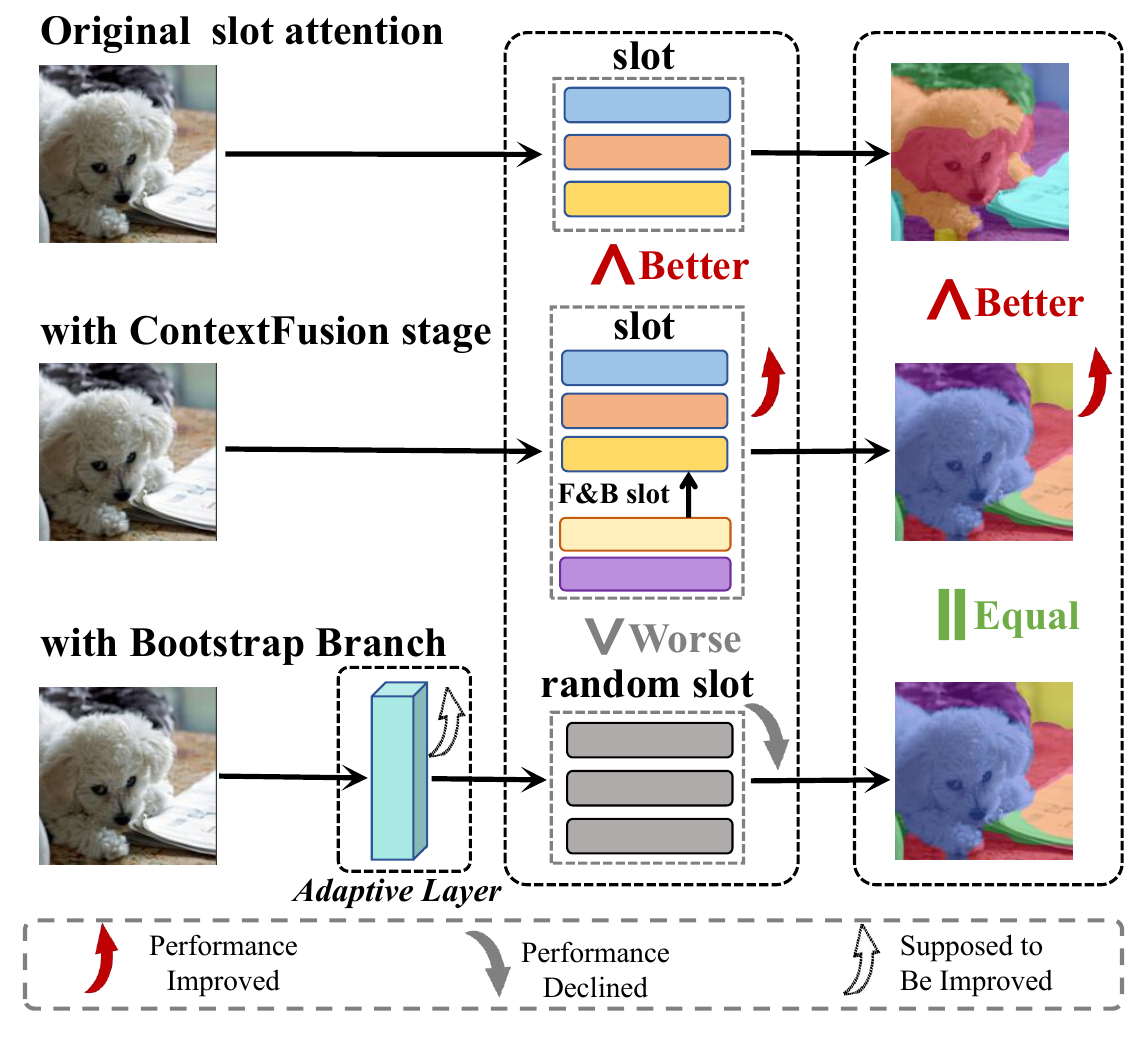}
    \caption{\textbf{Design Philosophy of the Bootstrap Branch.} In the Bootstrap Branch, the feature adapted by the adaptation layer is designed to match the superior results of the ContextFusion stage, with the inferior slot.}
    \label{fig:boostrap motivation}
\end{figure}

To ensure that the feature adaptation layer enhances the feature representation capability, we use a slot attention module, different from the one in the ContextFusion stage, to guide image region binding into the appropriate slots within the Bootstrap Branch. The slot attention module is initialized from scratch and then momently optimized by the slot attention module from the ContextFusion stage.

The rationale behind this design is that the adapted features, when combined with coarse slots, are expected to achieve the performance improvements observed in the ContextFusion stage. The coarse slot attention increases the difficulty of slot binding, and only a better feature representation can achieve competitive performance. The slot attention module is frozen when we use the matching loss to optimize the feature adaptation layer. Fig. \ref{fig:boostrap motivation} illustrates the motivation behind this design. Freezing the slot attention module during the matching loss optimization emphasizes the refinement of the feature adaptation layer.

At the testing stage, the feature-adaptive layer in the Bootstrap Branch and the latest slot attention module from the ContextFusion stage are combined to leverage their respective strengths, achieving the final performance. Although these components reside in separate branches, our design uses a feature adaptation layer to make slight adjustments to the encoder's features rather than fine-tuning the entire encoder. The Bootstrap Branch follows a similar pipeline, using the slot attention module to obtain results. These designs ensure that the features outputted by the adaptation layer in the Bootstrap Branch are compatible with the latest slot attention block with fusion layer in the ContextFusion stage.

\section{Experiments}\label{sec:experiments}

In our experiments, we adopt two different state-of-the-art slot attention models as subjects, integrating them with our method to observe changes in performance. Three benchmarks are used, including synthetic and natural images, to evaluate the performance across different settings.
% ——————————————————————————————————————————————————————————————————————————————————
\subsection{Setup}
\aboverulesep=0pt
\belowrulesep=0pt
\begin{table*}[!ht]
    \centering
    \small
    \caption{Comparison with different object-centric methods on MOVi-C, PASCAL, and COCO datasets.} 
    \renewcommand{\arraystretch}{1.2}
    \begin{tabular*}{1.01\textwidth}{
    >{\raggedleft\arraybackslash}m{4.8cm}|
    >{\centering\arraybackslash}m{1.4cm}>{\centering\arraybackslash}m{1.4cm}|
    >{\centering\arraybackslash}m{1.4cm}>{\centering\arraybackslash}m{1.4cm}|
    >{\centering\arraybackslash}m{1.4cm}>{\centering\arraybackslash}m{1.4cm}}
    \specialrule{1.5pt}{0pt}{0pt} 
    \multirow{2}{5cm}{\centering Method} & \multicolumn{2}{c}{MOVi-C} & \multicolumn{2}{c}{PASCAL} & \multicolumn{2}{c}{COCO}  \\ \cmidrule{2-7} 
        ~ & mBO$^{i}$ & mIoU & mBO$^{i}$ & mBO$^{c}$ & mBO$^{i}$ & mBO$^{c}$   \\ \hline
        Slot Attention \citep{SA}  \textcolor{gray}{\scriptsize{[NeurIPS 2020]}}          & 26.3 & - & 24.6 & 24.9 &  17.2 & 19.2 \\
        SLASH \citep{SLASH} \textcolor{gray}{\scriptsize{[CVPR 2023]}}  & - & 27.7 & - & - & - & -   \\
        SLATE \citep{SLATE} \textcolor{gray}{\scriptsize{[ICLR 2022]}}    & 39.4 & 37.8 & 35.9 & 41.5 & 29.1 & 33.6 \\
        SlotDiffusion \citep{SlotDiffusion} \textcolor{gray}{\scriptsize{[NeurIPS 2023]}}          & - & - & 50.4 & 55.3 & 31.0 & 35.0 \\
        (Stable-) LSD \citep{LSD} \textcolor{gray}{\scriptsize{[NeurIPS 2023]}}    & 45.6 & 44.2 & - & - & 30.4 & -  \\ 
        Top-Down Slot-Attention \citep{kim2024bootstrapping} \textcolor{gray}{\scriptsize{[NeurIPS 2024]}}          & 46.8 & 45.9 & 43.9 & 51.0 & \textbf{33.0} & 40.3  \\
        OCEBO \citep{OCEBO} \textcolor{gray}{\scriptsize{[ICLR 2025]}}          & 27.3 & - & 34.4 & - & - & - \\ 
        
        FT-DINOSAUR \citep{FT-DINOSAUR} \textcolor{gray}{\scriptsize{[arXiv 2024]}}          & 44.2 & - & 37.6 & - & - & - \\\specialrule{1.2pt}{0pt}{0pt}

        DINOSAUR  \citep{randomframes}  \textcolor{gray}{\scriptsize{[ICLR 2023]}}  & 42.4 & 41.8 & 44.0 & 51.2 & 31.1 & 38.9    \\ \rowcolor{gray!20}
        DINOSAUR + Ours  & \textbf{47.0} \hspace{0.01cm} \greenarrow  \hspace{0.01cm} \textcolor{green!50!black}{4.6} & \textbf{46.0} \hspace{0.01cm} \greenarrow  \hspace{0.01cm} \textcolor{green!50!black}{4.2} & \textbf{46.6} \hspace{0.01cm} \greenarrow  \hspace{0.01cm} \textcolor{green!50!black}{2.6} & \textbf{54.1} \hspace{0.01cm} \greenarrow  \hspace{0.01cm} \textcolor{green!50!black}{2.9} & 32.2 \hspace{0.01cm} \greenarrow  \hspace{0.01cm} \textcolor{green!50!black}{1.0} & \textbf{41.7} \hspace{0.01cm} \greenarrow  \hspace{0.01cm} \textcolor{green!50!black}{2.8} \\ \hline 
        SPOT \citep{spot}  \textcolor{gray}{\scriptsize{[CVPR 2024]}}  & 47.0 & 46.4 & 48.1 & 55.3 & 34.7 & 44.3 \\ \rowcolor{gray!20}
        
        SPOT + Ours    & \textbf{49.5} \hspace{0.01cm} \greenarrow  \hspace{0.01cm} \textcolor{green!50!black}{2.5} & \textbf{49.2} \hspace{0.01cm} \greenarrow  \hspace{0.01cm} \textcolor{green!50!black}{2.8} & \textbf{49.8} \hspace{0.01cm} \greenarrow  \hspace{0.01cm} \textcolor{green!50!black}{1.7} & \textbf{57.0} \hspace{0.01cm} \greenarrow  \hspace{0.01cm} \textcolor{green!50!black}{1.7} & \textbf{35.8} \hspace{0.01cm} \greenarrow  \hspace{0.01cm} \textcolor{green!50!black}{1.1} & \textbf{45.6} \hspace{0.01cm} \greenarrow  \hspace{0.01cm} \textcolor{green!50!black}{1.3} \\
        \specialrule{1.5pt}{0pt}{0pt}
    \end{tabular*}   

    \vspace{0.5em}
    \noindent\parbox{\linewidth}{
    \footnotesize
    DINOSAUR uses an autoregressive decoder and DINO \citep{DINO} ViT encoder. The experimental results of the baseline are derived from the original corresponding papers.
    }

    \label{tab:main}
\end{table*}

\textbf{Datasets.} 
We evaluate our method on the task of unsupervised object discovery using three datasets: one synthetic dataset, MOVi-C \citep{movic}, and two real-world datasets, PASCAL VOC 2012 \citep{voc} and MS COCO 2017 \citep{coco}. A brief introduction to each dataset is provided below:

\textit{MOVi-C} contains approximately $1,000$ real 3D-scanned objects, with each scene comprising $3$ to $10$ objects. Although originally developed as a video dataset, we adapt it for image-based evaluation by randomly sampling frames, following the approach in \citep{randomframes}.

\textit{PASCAL VOC 2012} is a real-world image dataset featuring natural scenes with diverse objects. Most scenes in this dataset typically contain only one or a few salient objects, making it a relatively straightforward benchmark for evaluation.

\textit{MS COCO 2017} is another real-world dataset that is widely used for object detection and segmentation tasks, similar to PASCAL VOC 2012. However, it features more complex scenes with multiple objects. Due to this complexity, MS COCO 2017 presents a significant challenge for object-centric learning models.

In this paper, we refer to the two real-world datasets as \textit{PASCAL} and \textit{COCO} for short.
    
\textbf{Metrics.} To comprehensively evaluate object-centric learning performance, we adopt two widely used metrics: Mean Best Overlap (mBO) and Mean Intersection over Union (mIoU), as suggested in \citep{spot}.

\textit{mBO} evaluates how well the predicted object masks align with the ground-truth masks. It includes two variants: instance-level mBO$^{i}$ and category-level mBO$^{c}$. Specifically, mBO$^{i}$ matches each predicted mask to the instance-level ground-truth masks, while mBO$^{c}$ performs the matching with semantic (category-level) masks. For each ground-truth object, the predicted mask with the highest Intersection over Union (IoU) is selected, and the final score is computed as the average IoU over all matched pairs.

\textit{mIoU} establishes a one-to-one correspondence between predicted and ground-truth masks using the Hungarian algorithm and computes the average IoU over all matched pairs. Unlike mBO, which selects the best-matching predicted mask for each ground-truth object individually, mIoU reflects the overall segmentation performance more comprehensively through a globally optimal matching strategy.

In contrast, although FG-ARI has been used in previous work as a clustering-based segmentation metric, it often fails to capture the actual mask accuracy and can lead to misleading evaluations \citep{spot}. Hence, FG-ARI is not reported in this paper.

% To assess object-centric learning, we used Mean Best Overlap (mBO) and Mean Intersection over Union (mIoU). The Mean Best Overlap (mBO) includes instance-level mBO$^{i}$ and category-level mBO$^{c}$. mBO$^{i}$ and mBO$^{c}$ match the predicted masks to the instance and semantic masks, respectively, by finding the best overlap. Meanwhile, mIoU employs the Hungarian algorithm to match predicted and ground truth masks. \citep{spot} suggests that FG-ARI, as a clustering similarity metric, may yield unreliable segmentation quality assessments by overlooking the accuracy of predicted masks. Therefore, FG-ARI is not reported in this paper.
    
\textbf{Implementation Details.} In this section, we provide implementation details for the auxiliary indicator, as well as the training parameters used in our ContextFusion Stage and Bootstrap Branch.

\textit{Auxiliary Indicator.} For the indicator, we use ViT-B/16 as the encoder (initialized with DINO by default) and optimize only the final layer. We employ the SGD optimizer with a learning rate set to $0.001$ and a batch size of $512$. The learning rate follows a linear warm-up for $10,000$ steps and then an exponential decay schedule. Additionally, we clip the gradient norm to $1$ to stabilize the training process. We train for $100$ epochs across all datasets.

For data augmentation in the indicator, we apply color jitter with a probability of $80\%$, convert the image to grayscale with a probability of $20\%$, flip the image horizontally with a probability of $50\%$, and invert the image pixel values with a probability of $20\%$. In the two branches of contrastive learning, Gaussian blur is applied with probabilities of $100\%$ and $10\%$, respectively, in the two contrastive branches..

For all datasets, we set the loss weights of $\mathcal{L}^\text{pixel}$, $\mathcal{L}^\text{stuff}$, and $\mathcal{L}^\text{sep}$ to $0.5$, $0.5$, and $0.5$, respectively.

\textit{ContextFusion Stage and Bootstrap Branch.} In both of them, we use ViT-B/16 as the encoder (initialized with DINO by default) and freeze it during training. We employ the Adam optimizer \citep{kingma2014adam} with a learning rate of $4e-4$ and a batch size of $64$. Following \citep{randomframes}, we use $7$, $6$, and $11$ slots for the COCO, PASCAL, and MOVi-C datasets, respectively. 

\subsection{Integrating into SOTA Object-centric Methods}
\textbf{Baselines.} We apply our approach to two slot attention methods, DINOSAUR \citep{randomframes}, SPOT \citep{spot}. 

\textit{DINOSAUR} \citep{randomframes} introduces an architecture that combines DINO's self-supervised feature reconstruction loss with the slot attention module, incorporating an inductive bias based on the homogeneity of features within objects. This approach bridges the gap between object-centric representation learning on synthetic and real-world datasets. It has gradually become a new paradigm in the field of object-centric learning.

\textit{SPOT} \citep{spot} proposes a two-stage strategy to improve slot generation and segmentation accuracy by distilling high-quality slot attention masks from the decoder to the encoder via self-training. Additionally, it enhances the autoregressive decoder through sequence permutations, which strengthens the role of slots in reconstruction and provides stronger supervisory signals.

Furthermore, we also list the results of several other object-centric methods, including \textit{Slot Attention} \citep{SA}, \textit{SLASH} \citep{SLASH}, \textit{SLATE} \citep{SLATE}, \textit{SlotDiffusion} \citep{SlotDiffusion}, and \textit{LSD} \citep{LSD}, \textit{etc.}.
    
\textbf{Main Results on Each Dataset.} Tab. \ref{tab:main} presents the main experimental results of our method on the COCO, PASCAL, and MOVi-C datasets. Our approach consistently improves the performance of both DINOSAUR and SPOT across all datasets. Notably, the performance gains on the synthetic MOVi-C dataset are significantly higher than those on two real-world datasets. This is primarily because MOVi-C features relatively homogeneous backgrounds with simple textures, allowing our foreground-background indicator to produce more accurate segmentation results. Consequently, this enables more effective scene decomposition during the fusion stage.
%In addition, the performance improvement on SPOT is smaller compared to that over DINOSAUR. This is because our Bootstrap Branch is designed to enhance the representational capacity of encoder, while the encoder of SPOT has already been fine-tuned via self-training in the original method. Therefore, the benefit of our approach on SPOT is limited. In contrast, for methods like DINOSAUR, whose encoder cannot be finetuned, our Bootstrap Branch significantly enhances their representational capacity, leading to significantly improved performance.

% ——————————————————————————————————————————————————————————————————————————————————
\begin{table}[!ht]
    \centering
    % \captionsetup{position=top}
    \small
     \caption{The effect of the foreground and background semantic information provided by different indicators on the final performance of DINOSAUR.}
    % \renewcommand{\arraystretch}{1.2}
    % \resizebox{1 \linewidth}{!}{
    
    \begin{tabular*}{\linewidth}{@{}
    >{\centering\arraybackslash}m{1.2cm} |
    m{1.65cm} |
    >{\centering\arraybackslash}m{0.8cm} |
    >{\centering\arraybackslash}m{0.8cm} |
    >{\centering\arraybackslash}m{1.2cm}
@{}}

    \specialrule{1.5pt}{0pt}{0pt} 
        Dataset & Indicator & IoU & mBO$^{i}$ & mIoU / mBO$^{c}$   \\ \hline 
        \multirow{5}{*}{MOVi-C} & - & - & 42.4 & 41.7   \\
        ~ & Random  & - & 19.2 & 18.0  \\ 
        ~ & DINOSAUR & 45.2 & 45.5 & 44.6    \\
        ~ & SPOT & 46.6 & 45.8 & 44.9    \\
        ~ & \cellcolor{gray!20}Ours & \cellcolor{gray!20} \textbf{47.4} &\cellcolor{gray!20}  \textbf{47.0} & \cellcolor{gray!20} \textbf{46.0}    \\
        \specialrule{1.5pt}{0pt}{0pt}
        \multirow{5}{*}{PASCAL} & - & - & 43.9 & 51.3   \\
        ~ & Random  & - & 24.0 & 27.0  \\ 
        ~ & DINOSAUR & 47.6 & 41.5 & 45.7    \\
        ~ & SPOT & 50.8 & 45.2 & 52.0    \\
        ~ & \cellcolor{gray!20}Ours & \cellcolor{gray!20} \textbf{52.6} &\cellcolor{gray!20}  \textbf{46.6} & \cellcolor{gray!20} \textbf{54.1}    \\
    \specialrule{1.5pt}{0pt}{0pt}

        \multirow{5}{*}{COCO} & - & - & 31.1 & 38.9   \\
        ~ & Random  & - & 19.4 & 24.9  \\ 
        ~ & DINOSAUR & 44.5 & 30.8 & 39.1    \\
        ~ & SPOT & 46.2 & 31.3 & 39.5    \\
        ~ & \cellcolor{gray!20}Ours & \cellcolor{gray!20} \textbf{48.4} &\cellcolor{gray!20}  \textbf{32.2} & \cellcolor{gray!20} \textbf{41.7}    \\
        \specialrule{1.5pt}{0pt}{0pt}
    \end{tabular*}
    % }
    \vspace{0.5em}
    \noindent\parbox{\linewidth}{
    \footnotesize
    DINOSAUR and SPOT refer to configurations where the number of slots is set to 2. IoU is the metric used to measure the performance of the indicator in segmenting foreground and background. mIoU and $\text{mBO}^c$ are reported to evaluate the final performance of DINOSAUR after integrating knowledge from different indicators on the MOVi-C dataset, and on the PASCAL and COCO datasets, respectively.
    }
    \label{tab:indicator}
\end{table}

\subsection{Ablation Study}
In this section, we analyze various aspects of our approach: the significance of the foreground and background indicators and the effectiveness of each component.

\textbf{(A) Is the extra foreground and background semantic information necessary for slot attention methods?} We conducted several experiments on a real-world dataset (PASCAL) and a synthetic dataset (MOVi-C) to validate this question. Tab. \ref{tab:indicator} shows the results. In this table, "Random" refers to using random noise as the input to the fusion module, "DINOSAUR" and "SPOT" refer to setting the number of slots to 2 to provide the foreground and background slots as semantic information, and "Ours" refers to using the foreground and background semantic representation from our indicator into fusion model. All the results are conducted without the bootstrap branch. We convert the ground truth masks into binary masks containing only foreground ($0$) and background ($1$), and apply the best overlap matching method to compute the IoU metric by matching the predicted mask with the ground truth mask to show the prediction region accuracy about the foreground and background.

According to the results, we can find that our indicator achieves the best performance in recognizing the foreground and background regions on all the datasets. Autoencoder-based methods like DINOSAUR and SPOT, although capable of dividing the scene into two parts, still struggle to achieve accurate foreground-background segmentation. 

We further observe that an improved foreground-background indicator can substantially enhance the performance of slot attention-based methods. By comparing the final performance of DINOSAUR integrated with different sources of additional foreground and background semantic information, we find that random noise significantly degrades performance. Both the standard slot-attention methods and our approach can provide semantic representations of foreground and background regions, thus resulting in improved performance. However, our proposed method achieves the most accurate foreground-background decomposition, leading to the largest performance improvements.

\begin{figure}[ht]
    \centering
    \includegraphics[width=1\linewidth]{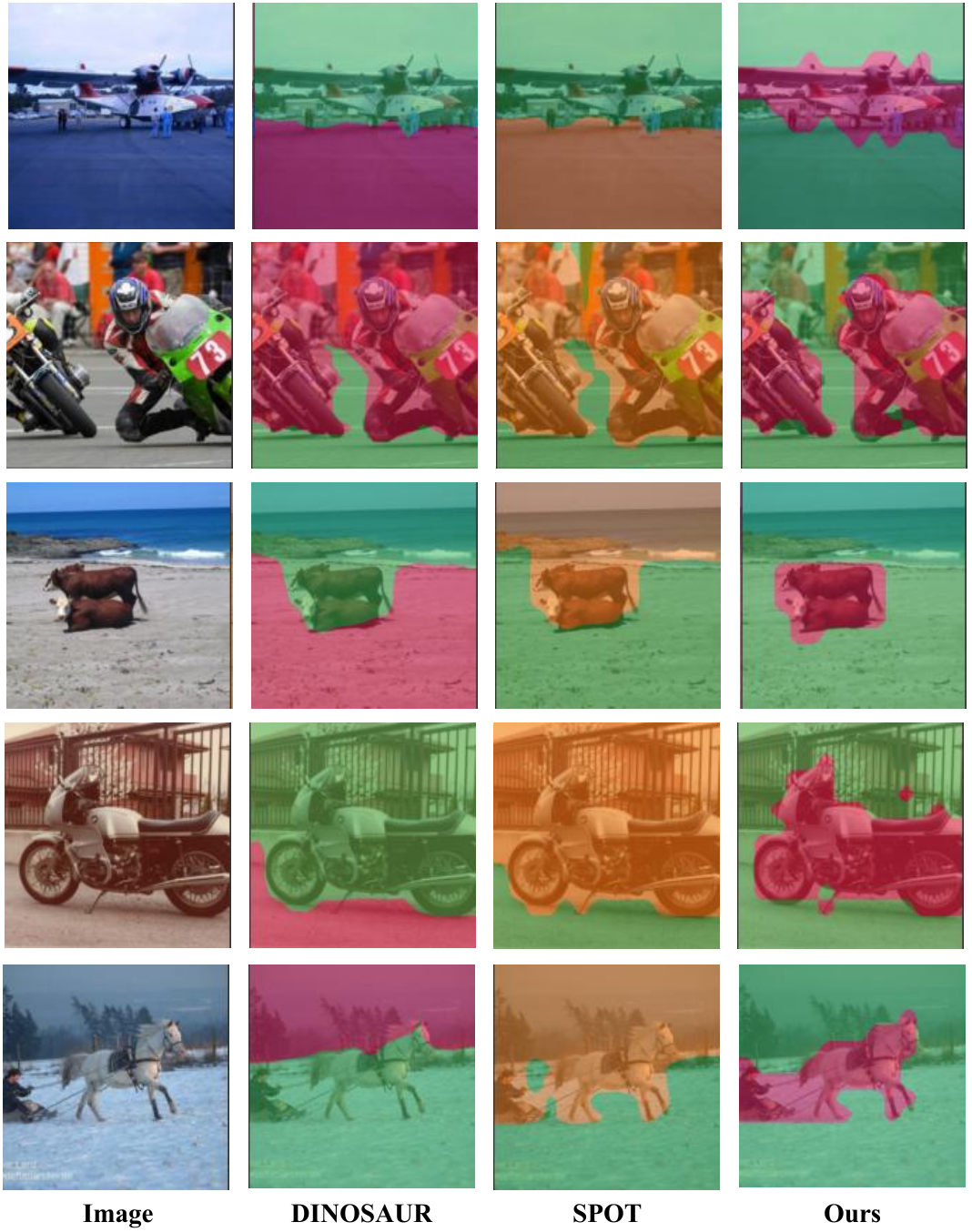}
    \caption{\textbf{Visualization of foreground and background regions provided by different indicators.} Compared to alternative indicators, our proposed method provides clearer and more accurate foreground-background distinctions.}
    \label{fig:bg}
\end{figure}

We illustrate several foreground-background region recognition results in Fig.~\ref{fig:bg}. The classical slot attention methods typically segment the scene into top and bottom regions without adequately considering semantic content. In contrast, our proposed method effectively distinguishes foreground and background regions based on semantics.

% Fusion_l
\aboverulesep=0pt
\belowrulesep=0pt
\begin{table*}[!ht]
    \centering
    \caption{Ablation analysis of individual components in our method.} 
    \renewcommand{\arraystretch}{1.2}
\small
    \begin{tabular*}{\textwidth}{@{}>{\centering\arraybackslash}m{1.65cm} |>{\centering\arraybackslash}m{0.9cm} >{\centering\arraybackslash}m{1.15cm} |
    >{\centering\arraybackslash}m{1.45cm} >{\centering\arraybackslash}m{1.45cm} |
    >{\centering\arraybackslash}m{1.45cm} >{\centering\arraybackslash}m{1.45cm}| 
    >{\centering\arraybackslash}m{1.45cm} >{\centering\arraybackslash}m{1.45cm} @{} }
    \specialrule{1.5pt}{0pt}{0pt}
        \multirow{2}{*}{Baseline} & \multicolumn{2}{c}{Module} & \multicolumn{2}{c}{MOVi-C} & \multicolumn{2}{c}{PASCAL} & \multicolumn{2}{c}{COCO}  \\ \cmidrule{2-9}

        ~ & \shortstack[b]{\rule{0pt}{0.8em}\footnotesize Context\\ \footnotesize Fusion} & \shortstack[b]{\rule{0pt}{0.8em}\footnotesize Bootstrap\\ \footnotesize Branch} & mBO$^{i}$ & mIoU & mBO$^{i}$ & mBO$^{c}$ & mBO$^{i}$ & mBO$^{c}$\\ [-2pt]
        \hline

        \multirow{3}{*}{DINOSAUR} & ~ & ~                & 42.4 & 41.8 & 44.0 & 51.2 & 31.1 & 38.9  \\
       ~   &\ding{51} & ~   & 45.8 \hspace{0.01cm} \greenarrow  \hspace{0.07cm} \textcolor{green!50!black}{3.4} & 44.7 \hspace{0.01cm} \greenarrow  \hspace{0.07cm} \textcolor{green!50!black}{2.9} & 45.9 \hspace{0.01cm} \greenarrow  \hspace{0.07cm} \textcolor{green!50!black}{1.9} & 52.8 \hspace{0.01cm} \greenarrow  \hspace{0.07cm} \textcolor{green!50!black}{1.6} & 31.7 \hspace{0.01cm} \greenarrow  \hspace{0.07cm} \textcolor{green!50!black}{0.6} & 40.6 \hspace{0.01cm} \greenarrow  \hspace{0.07cm} \textcolor{green!50!black}{1.7} \\ 

        % \rowcolor{gray!20}
        ~   &\cellcolor{gray!20}\ding{51} & \cellcolor{gray!20}\ding{51}   & \cellcolor{gray!20}\textbf{47.0} \hspace{0.01cm} \greenarrow  \hspace{0.07cm} \textcolor{green!50!black}{4.6} &\cellcolor{gray!20} \textbf{46.0} \hspace{0.01cm} \greenarrow  \hspace{0.07cm} \textcolor{green!50!black}{4.2} & \cellcolor{gray!20}\textbf{46.6} \hspace{0.01cm} \greenarrow  \hspace{0.07cm} \textcolor{green!50!black}{2.6} & \cellcolor{gray!20}\textbf{54.1} \hspace{0.01cm} \greenarrow  \hspace{0.07cm} \textcolor{green!50!black}{2.9} & \cellcolor{gray!20}\textbf{32.2} \hspace{0.01cm} \greenarrow  \hspace{0.07cm} \textcolor{green!50!black}{1.0} & \cellcolor{gray!20}\textbf{41.7} \hspace{0.01cm} \greenarrow  \hspace{0.07cm} \textcolor{green!50!black}{2.8} \\ \hline 
        
        \multirow{3}{*}{SPOT}& ~ & ~                   & 47.0 & 46.4 & 48.1 & 55.3 & 34.7 & 44.3   \\

        ~   &\ding{51} & ~  & 48.8 \hspace{0.01cm} \greenarrow  \hspace{0.07cm} \textcolor{green!50!black}{1.8} & 48.1 \hspace{0.01cm} \greenarrow  \hspace{0.07cm} \textcolor{green!50!black}{1.7}  & 49.2 \hspace{0.01cm} \greenarrow  \hspace{0.07cm} \textcolor{green!50!black}{1.1} & 56.3 \hspace{0.01cm} \greenarrow  \hspace{0.07cm} \textcolor{green!50!black}{1.0} & 35.2 \hspace{0.01cm} \greenarrow  \hspace{0.07cm} \textcolor{green!50!black}{0.5} & 45.2 \hspace{0.01cm} \greenarrow  \hspace{0.07cm} \textcolor{green!50!black}{0.9} \\
        
        % \rowcolor{gray!20}
         ~   &\cellcolor{gray!20}\ding{51} & \cellcolor{gray!20}\ding{51}    & \cellcolor{gray!20}\textbf{49.5} \hspace{0.01cm} \greenarrow  \hspace{0.07cm} \textcolor{green!50!black}{2.5} & \cellcolor{gray!20}\textbf{49.2} \hspace{0.01cm} \greenarrow  \hspace{0.07cm} \textcolor{green!50!black}{2.8} & \cellcolor{gray!20}\textbf{49.8} \hspace{0.01cm} \greenarrow  \hspace{0.07cm} \textcolor{green!50!black}{1.7} &\cellcolor{gray!20} \textbf{57.0} \hspace{0.01cm} \greenarrow  \hspace{0.07cm} \textcolor{green!50!black}{1.7} & \cellcolor{gray!20}\textbf{35.8} \hspace{0.01cm} \greenarrow  \hspace{0.07cm} \textcolor{green!50!black}{1.1} & \cellcolor{gray!20}\textbf{45.6} \hspace{0.01cm} \greenarrow  \hspace{0.07cm} \textcolor{green!50!black}{1.3} \\
        \specialrule{1.5pt}{0pt}{0pt}
    \end{tabular*}
% }

     \vspace{0.5em} % ← 控制表格与解释之间间距
    \noindent\parbox{\linewidth}{
    \footnotesize
    % \textit{Fusion Layer} refers to the basic slot fusion operation (Eq.\eqref{eq:fusion}), \textit{ContextFusion} denotes applying self-attention on the fused slots (Eq.\eqref{eq:sft}), and \textit{Ours} represents the full model with the bootstrap branch added to ContextFusion.
    }
    
    \label{tab:module_ablation}
\end{table*}

\begin{figure*}[ht]
    \centering
    \includegraphics[width=1\linewidth]{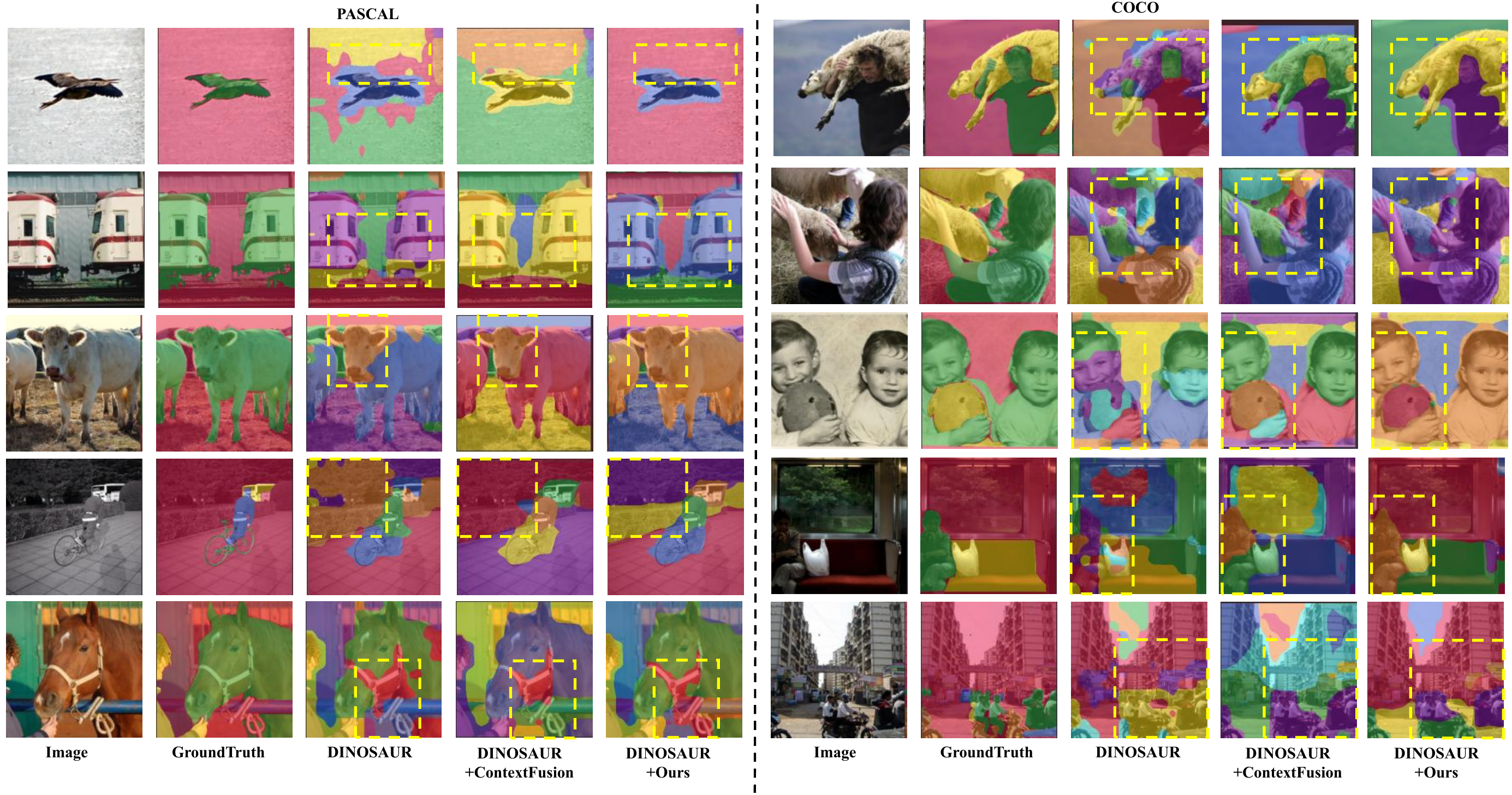}
    \caption{\textbf{Visualization results of different components within our method.} `ContextFusion' refers to the use of only the ContextFusion stage, whereas `Ours' represents the complete version of our method.
}
    \label{fig:component}
\end{figure*}

\textbf{(B) The effectiveness of each component in our method.} Tab. \ref{tab:module_ablation} presents the performance of our method with or not each proposed component on the MOVi-C, PASCAL, and COCO datasets.
The results indicate that the ContextFusion module improves predictions by fusing the extra semantic information into existing methods through the fusion layer. Based on the improved outputs from the ContextFusion stage, our bootstrap branch utilizes these outputs as targets to further boost the final performance. Fig. \ref{fig:component} presents the visualization results of different components within our method. As demonstrated, the ContextFusion stage enables the model to discern which parts may belong to the same foreground object and to unite them. Meanwhile, the Bootstrap Branch further enhances the outcomes of the ContextFusion stage, particularly around the edges of objects.

%=============================================================
\subsection{More Discussion}

\textbf{(A) Impact of different losses in foreground and background indicator.} We conduct ablation studies to investigate the role of different loss functions in the foreground-background indicator (Tab.~\ref{tab:loss}). The results show that the separation loss \( \mathcal{L}^{\text{sep}} \) plays a pivotal role. When only the pixel-level loss \( \mathcal{L}^{\text{pixel}} \) and stuff-level loss \( \mathcal{L}^{\text{stuff}} \) are used, the model tends to collapse into a degenerate solution in which all visual information is compressed into a single slot (one slot passively responds to the entire image), while the other remains completely inactive. Although this degenerate configuration yields low loss values for \( \mathcal{L}^{\text{pixel}} \) and \( \mathcal{L}^{\text{stuff}} \), it fundamentally fails to perform meaningful foreground-background separation.

\begin{table}%[!ht]
    \centering
    \caption{Ablation study of different loss functions for our auxiliary indicator.}
    \renewcommand{\arraystretch}{1.5}
    
    % \resizebox{0.95\linewidth}{!}{
    {\scriptsize
    \begin{tabular*}{\linewidth}{c c c|>{\centering\arraybackslash}m{1cm}| c c}
    \specialrule{1.5pt}{0pt}{0pt}
        \( \mathcal{L}^{\text{pixel}} \) & \( \mathcal{L}^{\text{stuff}} \) & \( \mathcal{L}^{\text{sep}} \) & IoU & mBO$^{c}$ & mBO$^{i}$    \\ \hline
        \ding{51} &   &   & 28.8 & 52.2 & 45.2    \\ 
        \ding{51} & \ding{51} &  & 28.4 & 52.3 & 45.2   \\ 
        \ding{51} &  & \ding{51} & 48.9 & 52.9 & 45.8   \\ \rowcolor{gray!20}
        \ding{51} & \ding{51} & \ding{51} & \textbf{52.6} & \textbf{54.1} & \textbf{46.6}    \\
        \specialrule{1.5pt}{0pt}{0pt}
    \end{tabular*}
    }
    
    \vspace{0.5em} % ← 控制表格与解释之间间距
    \noindent\parbox{\linewidth}{
    \footnotesize
    IoU measures the indicator’s performance in segmenting foreground and background regions. $\text{mBO}^{c}$ and $\text{mBO}^{i}$ indicate the final performance of DINOSAUR after integration of the semantic information provided by these different indicators. All the experiments are conducted on the PASCAL.
    }
     
    \label{tab:loss}
\end{table}

To address this issue, our proposed separation loss \( \mathcal{L}^{\text{sep}} \) encourages the semantic centers to attend to distinct visual content, effectively promoting feature diversity. However, further experiments reveal that relying solely on \( \mathcal{L}^{\text{sep}} \) and \( \mathcal{L}^{\text{pixel}} \) renders the model overly sensitive to background variations. Only when \( \mathcal{L}^{\text{sep}} \), \( \mathcal{L}^{\text{pixel}} \), and \( \mathcal{L}^{\text{stuff}} \) are integrated cohesively can the model achieve stable and high-quality foreground-background disentanglement.

Notably, our quantitative analysis shows a positive correlation between the indicator's performance and the final model results. This finding further supports that the higher quality of foreground and background semantic information can better enhance the slot attention based methods.

% \begin{table}%[!ht]
%     \centering
%     \caption{Ablation study of different loss functions for our auxiliary indicator.}
%     \renewcommand{\arraystretch}{1.2}
%     \resizebox{0.85\linewidth}{!}{
%     \begin{tabular}{c c c|c| c c}
%     \specialrule{1.5pt}{0pt}{0pt}
%         \( \mathcal{L}^{\text{pixel}} \) & \( \mathcal{L}^{\text{stuff}} \) & \( \mathcal{L}^{\text{sep}} \) & IoU & mBO$^{c}$ & mBO$^{i}$    \\ \hline
%         \ding{51} &   &   & 28.8 & 52.2 & 45.2    \\ 
%         \ding{51} & \ding{51} &  & 28.4 & 52.3 & 45.2   \\ 
%         \ding{51} &  & \ding{51} & 48.9 & 52.9 & 45.8   \\ \rowcolor{gray!20}
%         \ding{51} & \ding{51} & \ding{51} & \textbf{52.6} & \textbf{54.1} & \textbf{46.6}    \\
%         \specialrule{1.5pt}{0pt}{0pt}
%     \end{tabular}
%     }
    
%     \vspace{0.5em} % ← 控制表格与解释之间间距
%     \noindent\parbox{\linewidth}{
%     \footnotesize
%     IoU measures the indicator’s performance in segmenting foreground and background regions. $\text{mBO}^{c}$ and $\text{mBO}^{i}$ indicate the final performance of DINOSAUR after integration of the semantic information provided by these different indicators. All the experiments are conducted on the PASCAL.
%     }
     
%     \label{tab:loss}
% \end{table}

%=============================================================
\textbf{(B) Impact of different pretrained models.} The capability of the encoder is crucial to the performance of slot attention methods. In Table~\ref{tab:DINOV2}, we adopt different pretrained models as the encoder (DINOv2~\citep{dinov2} and MAE~\citep{MAE}), investigating the robustness of our method across various encoders. %All experiments are conducted on the PASCAL dataset, and the results are shown in Table~\ref{tab:DINOV2}. Specifically, “DINO” refers to ViT-B/16 pretrained with DINO, “DINOv2” refers to ViT-B/14 pretrained with DINOv2, and “MAE” refers to ViT-B/16 pretrained with Masked Autoencoders.

The results demonstrate that our proposed ContextFusion and Bootstrap Branch modules consistently enhance performance across all encoder types. Among these encoders, the original DINO encoder achieves the best performance, although DINOv2 is regarded as an improved version of DINO.

% \begin{table}[!ht]
%     \centering
%     \caption{Impact of different pretrained models. This experiment is conducted on the PASCAL dataset. "DINO" refers to ViT-B/16 initialized with DINO, "DINOv2" refers to ViT-B/14 initialized with DINOv2, while "MAE" refers to ViT-B/16 initialized with MAE.}
%     \renewcommand{\arraystretch}{1.2}
%     % \resizebox{1\linewidth}{!}{
%     \begin{tabular}{c | l|c c }
%     \specialrule{1.5pt}{0pt}{0pt} 
%         Encoder & Methods &  mBO$^{i}$ & mBO$^{c}$   \\ \hline 
%         \multirow{3}{*}{DINO} & DINOSAUR & 44.0 & 51.2   \\
%         ~ & DINOSAUR + ContextFusion  & 45.9 & 52.8   \\ 
%         ~ & \cellcolor{gray!20}DINOSAUR + Ours & \cellcolor{gray!20}\textbf{46.6} & \cellcolor{gray!20}\textbf{54.1}     \\
%         \specialrule{1.5pt}{0pt}{0pt}
%         \multirow{3}{*}{DINOv2} & DINOSAUR & 40.7 & 47.6   \\
%         ~ & DINOSAUR + ContextFusion  & 42.0 & 48.8   \\ 
%         ~ & \cellcolor{gray!20}DINOSAUR + Ours & \cellcolor{gray!20}\textbf{42.9} & \cellcolor{gray!20}\textbf{49.9}     \\
%         \specialrule{1.5pt}{0pt}{0pt}
%         \multirow{3}{*}{MAE} & DINOSAUR & 38.0 & 44.3   \\
%         ~ & DINOSAUR + ContextFusion  & 39.2 & 45.3   \\ 
%         ~ & \cellcolor{gray!20}DINOSAUR + Ours & \cellcolor{gray!20}\textbf{40.0} & \cellcolor{gray!20}\textbf{46.0}     \\
%         \specialrule{1.5pt}{0pt}{0pt}
%     \end{tabular}
%     % }
%     \label{tab:DINOV2}
% \end{table}

\begin{table}[!ht]
    \centering
    \caption{Impact of different pretrained models. This experiment is conducted on the PASCAL dataset. "DINO" refers to ViT-B/16 initialized with DINO, "DINOv2" refers to ViT-B/14 initialized with DINOv2, while "MAE" refers to ViT-B/16 initialized with MAE.}
    \renewcommand{\arraystretch}{1.2}
    \begin{tabularx}{\linewidth}{@{}m{1cm} m{3.9cm}XX}
     \specialrule{1.5pt}{0pt}{0pt} 
        Encoder & Methods &  mBO$^{i}$ & mBO$^{c}$   \\ \hline 
        \multirow{3}{*}{DINO} & DINOSAUR & 44.0 & 51.2   \\
        ~ & DINOSAUR + ContextFusion  & 45.9 & 52.8   \\ 
        ~ & \cellcolor{gray!20}DINOSAUR + Ours & \cellcolor{gray!20}\textbf{46.6} & \cellcolor{gray!20}\textbf{54.1}     \\
        \specialrule{1.5pt}{0pt}{0pt}
        \multirow{3}{*}{DINOv2} & DINOSAUR & 40.7 & 47.6   \\
        ~ & DINOSAUR + ContextFusion  & 42.0 & 48.8   \\ 
        ~ & \cellcolor{gray!20}DINOSAUR + Ours & \cellcolor{gray!20}\textbf{42.9} & \cellcolor{gray!20}\textbf{49.9}     \\
        \specialrule{1.5pt}{0pt}{0pt}
        \multirow{3}{*}{MAE} & DINOSAUR & 38.0 & 44.3   \\
        ~ & DINOSAUR + ContextFusion  & 39.2 & 45.3   \\ 
        ~ & \cellcolor{gray!20}DINOSAUR + Ours & \cellcolor{gray!20}\textbf{40.0} & \cellcolor{gray!20}\textbf{46.0}     \\
        \specialrule{1.5pt}{0pt}{0pt}
    \end{tabularx}
    \label{tab:DINOV2}
\end{table}

\aboverulesep=0pt
\belowrulesep=0pt
\begin{table*}[!ht]
    \centering
    \caption{Roles of Cross-Attention and Self-Attention designs in our fusion layer.} 
    \renewcommand{\arraystretch}{1.2}
% \resizebox{0.9\textwidth}{!}{
    \small
    \begin{tabular*}{\textwidth}{@{}>{\centering\arraybackslash}m{1.65cm} |>{\centering\arraybackslash}m{1.15cm} >{\centering\arraybackslash}m{1.2cm} |
    >{\centering\arraybackslash}m{1.4cm} >{\centering\arraybackslash}m{1.4cm} |
    >{\centering\arraybackslash}m{1.4cm} >{\centering\arraybackslash}m{1.4cm}| 
    >{\centering\arraybackslash}m{1.4cm} >{\centering\arraybackslash}m{1.4cm} @{} }
    \specialrule{1.5pt}{0pt}{0pt}
                \multirow{2}{*}{Baseline}&\multicolumn{2}{c}{ContextFusion} & \multicolumn{2}{c}{MOVi-C} & \multicolumn{2}{c}{PASCAL} & \multicolumn{2}{c}{COCO}  \\ \cmidrule{2-9} 
        
        ~ & \shortstack[c]{\rule{0pt}{0.8em}\small Cross\\ \small Attention} & \shortstack[c]{\rule{0pt}{0.8em}\small Self\\ \small Attention} & mBO$^{i}$ &  mIoU & mBO$^{i}$ & mBO$^{c}$ & mBO$^{i}$ & mBO$^{c}$\\ [-2pt]\hline

        \multirow{3}{*}{DINOSAUR} & ~ & ~                & 42.4 & 41.8 & 44.0 & 51.2 & 31.1 & 38.9  \\
        
        ~ &\ding{51} & ~   & 45.0 \hspace{0.01cm} \greenarrow  \hspace{0.07cm} \textcolor{green!50!black}{2.6} & 44.2 \hspace{0.01cm} \greenarrow  \hspace{0.07cm} \textcolor{green!50!black}{2.4} & 45.5 \hspace{0.01cm} \greenarrow  \hspace{0.07cm} \textcolor{green!50!black}{1.5} & 52.3 \hspace{0.01cm} \greenarrow  \hspace{0.07cm} \textcolor{green!50!black}{1.1} &  31.9 \hspace{0.01cm} \greenarrow  \hspace{0.07cm} \textcolor{green!50!black}{0.8} & 40.4 \hspace{0.01cm} \greenarrow  \hspace{0.07cm} \textcolor{green!50!black}{1.5}  \\

       ~&\cellcolor{gray!20}\ding{51} & \cellcolor{gray!20}\ding{51}   & \cellcolor{gray!20}45.8 \hspace{0.01cm} \greenarrow  \hspace{0.07cm} \textcolor{green!50!black}{3.4} & \cellcolor{gray!20}44.7 \hspace{0.01cm} \greenarrow  \hspace{0.07cm} \textcolor{green!50!black}{2.9} & \cellcolor{gray!20}45.9 \hspace{0.01cm} \greenarrow  \hspace{0.07cm} \textcolor{green!50!black}{1.9} & \cellcolor{gray!20}52.8 \hspace{0.01cm} \greenarrow  \hspace{0.07cm} \textcolor{green!50!black}{1.6} & \cellcolor{gray!20}31.7 \hspace{0.01cm} \greenarrow  \hspace{0.07cm} \textcolor{green!50!black}{0.6} & \cellcolor{gray!20}40.6 \hspace{0.01cm} \greenarrow  \hspace{0.07cm} \textcolor{green!50!black}{1.7} \\ \hline

        \multirow{3}{*}{SPOT} & ~ & ~                   & 47.0 & 46.4 & 48.1 & 55.3 & 34.7 & 44.3   \\
        ~&\ding{51} & ~  & 48.2 \hspace{0.01cm} \greenarrow  \hspace{0.07cm} \textcolor{green!50!black}{1.2} & 47.8 \hspace{0.01cm} \greenarrow  \hspace{0.07cm} \textcolor{green!50!black}{1.4} & 49.0 \hspace{0.01cm} \greenarrow  \hspace{0.07cm} \textcolor{green!50!black}{0.9} & 56.3 \hspace{0.01cm} \greenarrow  \hspace{0.07cm} \textcolor{green!50!black}{1.0} & 35.2 \hspace{0.01cm} \greenarrow  \hspace{0.07cm} \textcolor{green!50!black}{0.5} & 45.0 \hspace{0.01cm} \greenarrow  \hspace{0.07cm} \textcolor{green!50!black}{0.7}  \\
        
        ~ &\cellcolor{gray!20}\ding{51} & \cellcolor{gray!20}\ding{51}  & \cellcolor{gray!20}48.8 \hspace{0.01cm} \greenarrow  \hspace{0.07cm} \textcolor{green!50!black}{1.8} & \cellcolor{gray!20}48.1 \hspace{0.01cm} \greenarrow  \hspace{0.07cm} \textcolor{green!50!black}{1.7}  & \cellcolor{gray!20}49.2 \hspace{0.01cm} \greenarrow  \hspace{0.07cm} \textcolor{green!50!black}{1.1} & \cellcolor{gray!20}56.3 \hspace{0.01cm} \greenarrow  \hspace{0.07cm} \textcolor{green!50!black}{1.0} & \cellcolor{gray!20}35.2 \hspace{0.01cm} \greenarrow  \hspace{0.07cm} \textcolor{green!50!black}{0.5} &\cellcolor{gray!20} 45.2 \hspace{0.01cm} \greenarrow  \hspace{0.07cm} \textcolor{green!50!black}{0.9} \\
        \specialrule{1.5pt}{0pt}{0pt}
    \end{tabular*}
    % }
    %\vspace{1em} % ← 控制表格与解释之间间距
    \noindent\parbox{\linewidth}{
    \footnotesize
    \vspace{1mm}
    Cross Attention refers to Eq.~\ref{eq:fusion}, and Self Attention refers to Eq.~\ref{eq:sft}.
    }

    \label{tab:attention_ablation}
\end{table*}

 \textbf{(C) The structure of the fusion layer.} Our proposed fusion layer comprises two sub-modules: Cross-Attention and Self-Attention. The Cross-Attention module is designed to effectively integrate foreground and background information into each slot. The Self-Attention module enhances the cooperation between slots. In this section, we investigate the impact of these two sub-modules. All experiments are conducted using only the ContextFusion stage, and the results are presented in Tab.~\ref{tab:attention_ablation}. These results suggest that the Cross-Attention module contributes more substantially within the fusion layer, as it primarily injects foreground-background information into slots. Meanwhile, the Self-Attention module serves an auxiliary role, refining slot representations.

%=============================================================

\textbf{(D) Consistent improvement across different slot numbers.} The number of slots used during model training can significantly impact the quality of the final results. Therefore, evaluating the model's robustness to variations in slot numbers is essential. In this section, we compare our proposed method with DINOSAUR to assess its stability under different slot configurations. We conduct experiments on one real-world dataset (PASCAL) and one synthetic dataset (MOVi-C), using five different slot settings. As shown in Tab. \ref{tab:slot_num}, our method consistently outperforms DINOSAUR across all slot numbers and exhibits lower performance variance.

\aboverulesep=0pt
\belowrulesep=0pt
\begin{table}[!ht]
    \centering
    \caption{The performance of our method with different numbers of slot.}
    
    \renewcommand{\arraystretch}{1.2}
    % \resizebox{0.96\linewidth}{!}{
    \begin{tabular*}{1.0\linewidth}{@{}>{\centering\arraybackslash}m{1.1cm} 
    |
    c | 
    >{\centering\arraybackslash}m{0.6cm} 
    >{\centering\arraybackslash}m{0.7cm} | 
    >{\centering\arraybackslash}m{1.34cm}
    >{\centering\arraybackslash}m{1.34cm}@{}}
     \specialrule{1.5pt}{0pt}{0pt} 
     \multirow{2}{*}{Dataset} &\multirow{2}{*}{$K$}& \multicolumn{2}{c}{DINOSAUR} & \multicolumn{2}{c}{DINOSAUR + Ours} \\ \cmidrule{3-4} \cmidrule{5-6} 
        ~ & ~ & mBO$^{i}$ & mBO$^{c}$  & mBO$^{i}$ & mBO$^{c}$ \\ \hline
        
        \multirow{5}{*}{PASCAL} & 2 & 34.3  & 40.4 & 38.3  \hspace{0.01cm} \greenarrow  \hspace{0.07cm} \textcolor{green!50!black}{4.0} & 45.9 \hspace{0.01cm} \greenarrow  \hspace{0.07cm} \textcolor{green!50!black}{5.5} \\ 
        ~ & 4 & 44.3  & 51.6  & 46.0 \hspace{0.01cm} \greenarrow  \hspace{0.07cm} \textcolor{green!50!black}{1.7} & 53.7  \hspace{0.01cm} \greenarrow  \hspace{0.07cm} \textcolor{green!50!black}{2.1}\\
        ~ & 6 & 44.8  & \textbf{51.8}  & 46.6 \hspace{0.01cm} \greenarrow  \hspace{0.07cm} \textcolor{green!50!black}{1.8} & \textbf{54.1} \hspace{0.01cm} \greenarrow  \hspace{0.07cm} \textcolor{green!50!black}{2.3} \\ 
        ~ & 8 & 45.3  & 51.7 & 46.7 \hspace{0.01cm} \greenarrow  \hspace{0.07cm} \textcolor{green!50!black}{1.4} & 54.0 \hspace{0.01cm} \greenarrow  \hspace{0.07cm} \textcolor{green!50!black}{2.3} \\ 
        ~ & 10 & \textbf{45.4} & 51.3  & \textbf{46.3} \hspace{0.01cm} \greenarrow \hspace{0.07cm} \textcolor{green!50!black}{0.9}  & 53.4  \hspace{0.01cm} \greenarrow \hspace{0.07cm} \textcolor{green!50!black}{2.1} \\ \hline

        \multirow{5}{*}{MOVi-C} & 2 & 13.8  & 9.0 & 16.0  \hspace{0.01cm} \greenarrow  \hspace{0.07cm} \textcolor{green!50!black}{2.2} & 11.5 \hspace{0.01cm} \greenarrow  \hspace{0.07cm} \textcolor{green!50!black}{2.5} \\ 
        ~ & 5 & 34.5  & 31.0  & 40.5 \hspace{0.01cm} \greenarrow  \hspace{0.07cm} \textcolor{green!50!black}{5.5} & 36.5  \hspace{0.01cm} \greenarrow  \hspace{0.07cm} \textcolor{green!50!black}{5.5}\\
        ~ & 7 & 39.8  & \textbf{37.8}  & 44.5 \hspace{0.01cm} \greenarrow  \hspace{0.07cm} \textcolor{green!50!black}{4.7} & \textbf{42.5} \hspace{0.01cm} \greenarrow  \hspace{0.07cm} \textcolor{green!50!black}{4.7} \\ 
        ~ & 9 & 41.9  & 40.7 & 46.5 \hspace{0.01cm} \greenarrow  \hspace{0.07cm} \textcolor{green!50!black}{4.6} & 45.2 \hspace{0.01cm} \greenarrow  \hspace{0.07cm} \textcolor{green!50!black}{4.5} \\ 
        ~ & 11 & \textbf{42.4} & 41.8  & \textbf{47.0} \hspace{0.01cm} \greenarrow \hspace{0.07cm} \textcolor{green!50!black}{4.6}  & 46.0  \hspace{0.01cm} \greenarrow \hspace{0.07cm} \textcolor{green!50!black}{4.2} \\
        \specialrule{1.5pt}{0pt}{0pt}
    \end{tabular*}
    % }

    \label{tab:slot_num}
\end{table}

\begin{figure*}[t]
    \centering
    \includegraphics[width=1\linewidth]{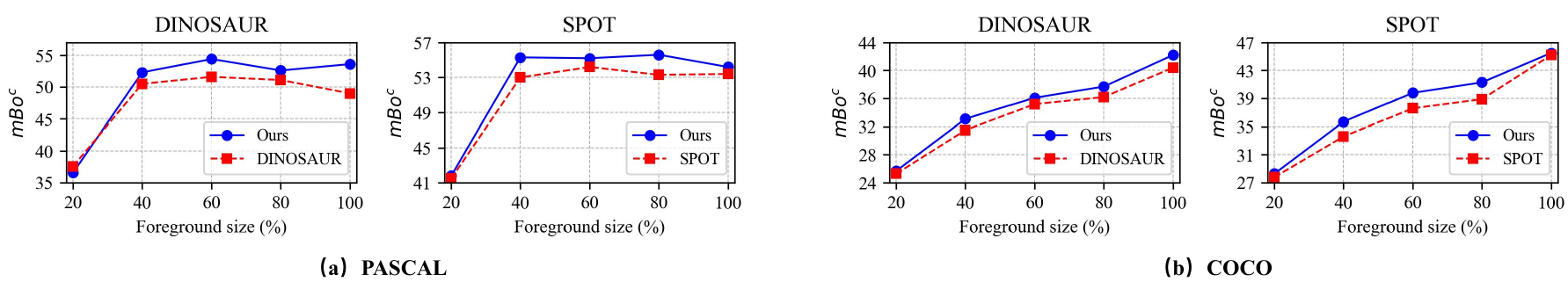}
    \caption{\textbf{Results across different foreground sizes}. All the experiments are conducted on COCO and PASCAL datasets, based on DINOSAUR and SPOT methods.}
    \label{fig:size}
\end{figure*}

\textbf{(E) Performance across different foreground sizes}.
Fig.~\ref{fig:size} presents the results obtained by DINOSAUR and SPOT, both with and without our method, on the PASCAL and COCO datasets across varying foreground object sizes. We calculate the proportion of foreground pixels relative to the total number of pixels and use this proportion to categorize the results. Experimental results demonstrate that our method consistently and significantly outperforms baseline methods over different foreground sizes. However, our method appears to struggle in scenes with smaller foreground objects.

\textbf{(F) Additional Training Time of Our Method.} 
The additional computational cost introduced by our method primarily includes two parts:
First, training the foreground-background indicator. Due to its simple architecture, training this indicator on the PASCAL dataset requires approximately 3 hours using a single A6000 GPU. Once trained, the indicator can be directly applied to different slot attention-based methods on the same dataset without retraining.
Second, when adapting existing slot attention-based methods into our ContextFusion and Bootstrap Branch framework, the fusion layer and adaptive layer only require training for a few epochs due to their simplicity.

% ——————————————————————————————————————————————————————————————————————————————————

% \begin{figure*}
%     \centering
%     \includegraphics[width=1\linewidth]{figure/component-sample.pdf}
%     \caption{\textbf{Visualization results of our method under different component configurations.} ``ContextFusion" indicates results obtained using only the ContextFusion stage. ``Ours" denotes the complete pipeline of the proposed method including the Bootstrap Branch.
% }
%     \label{fig:component}
% \end{figure*}

\begin{figure*}[t]
    \centering
    \includegraphics[width=1\linewidth]{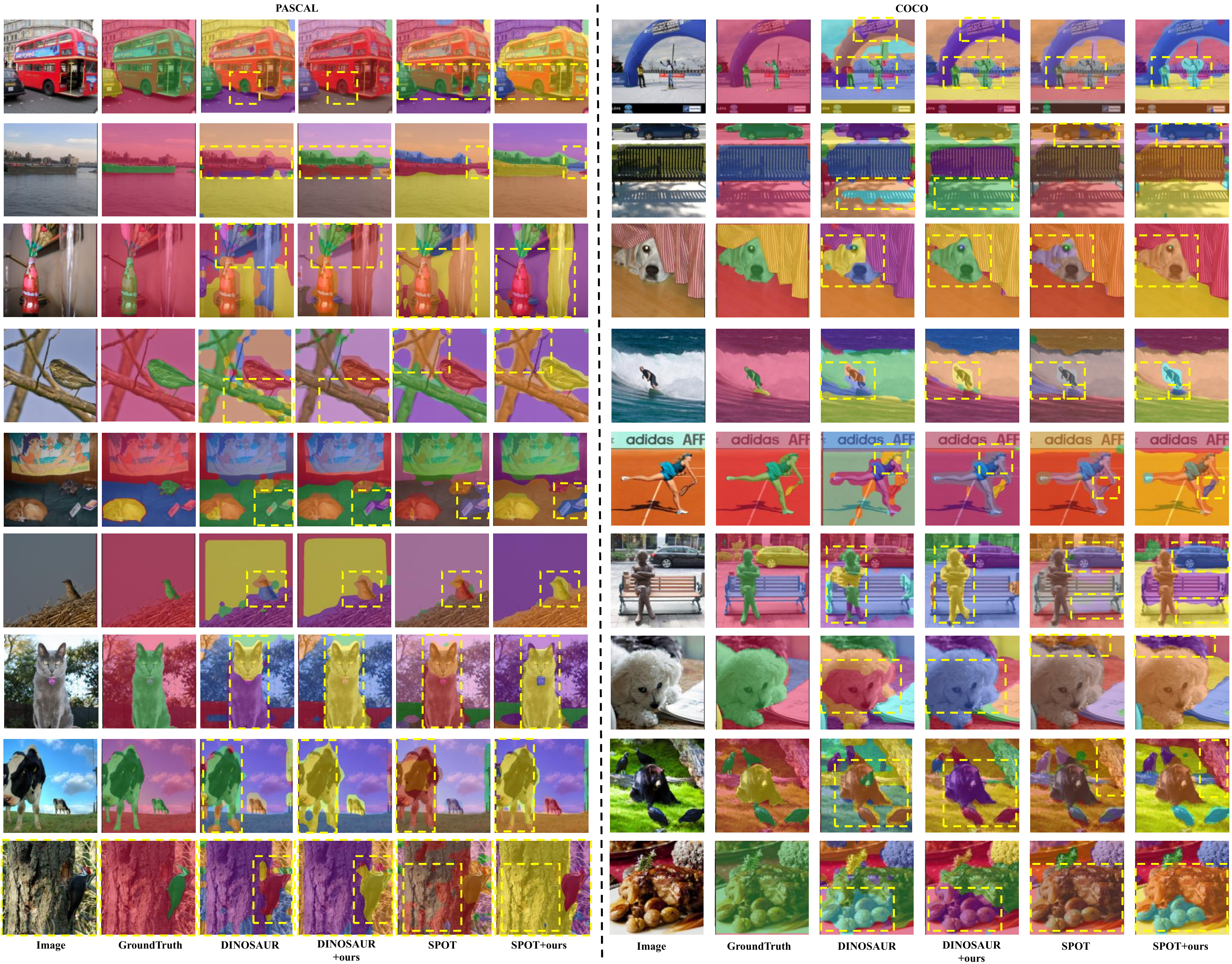}
    \caption{\textbf{Qualitative results on COCO and PASCAL datasets}.}
    \label{fig:main}
\end{figure*}

\subsection{Qualitative Results}
Fig. \ref{fig:main} presents qualitative results on the PASCAL and COCO datasets, comparing the performance of DINOSAUR and SPOT with and without our proposed method. The visualizations illustrate that by integrating additional semantic information, our method enhances the ability of existing techniques to more effectively identify elements in both the foreground and background. Consequently, objects in the foreground exhibit more complete outlines and more accurate identification results in diverse scenes.

\section{Conclusion}\label{sec:conclusion}

In this paper, we identify two key limitations in existing slot attention-based frameworks for object-centric learning: the lack of high-level semantic information and the inability to fine-tune the encoder. To address these issues, we propose a ContextFusion stage and a Bootstrap Branch. In the ContextFusion stage, we introduce an additional foreground and background indicator to provide extra semantic information, along with an attention fusion layer that integrates this knowledge into the slot attention-based method. In the Bootstrap Branch, we decouple feature adaptation from reconstruction and use the prediction results from the ContextFusion stage as the target for bootstrapping the training of the feature adaptation layer. At the testing stage, the feature adaptation layer, combined with the extra semantic information, is designed to enhance the performance of slot attention-based models. In our experiments, we integrate our method into three state-of-the-art slot attention models. The results consistently demonstrate that our method enhances performance across different datasets. Furthermore, it is important to mention that this study utilizes only foreground and background semantic information, without incorporating instance-level knowledge. We believe implementing coarse instance-level semantic information could potentially make significant advancements in our method.

\section*{Declarations}

\begin{itemize}
\item Funding. This work was supported by projects of the National Natural Science Foundation of China (Grant No.62206166, No.62302287), Shanghai Sailing Program (Grant No.23YF1413000), Shanghai Committee of Science and Technology (Grant No.23ZR1423500).
\item Data availability. The data that support the results and analysis of this
study is publicly available on their official network.
\end{itemize}
%\bmhead{Acknowledgements}

%Acknowledgements are not compulsory. Where included they should be brief. Grant or contribution numbers may be acknowledged.

%Please refer to Journal-level guidance for any specific requirements.

%%===========================================================================================%%
%% If you are submitting to one of the Nature Portfolio journals, using the eJP submission   %%
%% system, please include the references within the manuscript file itself. You may do this  %%
%% by copying the reference list from your .bbl file, paste it into the main manuscript .tex %%
%% file, and delete the associated \verb+\bibliography+ commands.                            %%
%%===========================================================================================%%

\bibliography{sn-bibliography}% common bib file
%% if required, the content of .bbl file can be included here once bbl is generated
%%\input sn-article.bbl

\end{document}